\documentclass[10pt, conference, letterpaper]{IEEEtran}
\IEEEoverridecommandlockouts
\usepackage{cite}
\usepackage{amsmath,amssymb,amsfonts}
\usepackage{algorithmic}
\usepackage{graphicx}
\usepackage{textcomp}
\usepackage{xcolor}
\def\BibTeX{{\rm B\kern-.05em{\sc i\kern-.025em b}\kern-.08em
    T\kern-.1667em\lower.7ex\hbox{E}\kern-.125emX}}
\usepackage{algorithm,algorithmic}
\usepackage{amsthm}
\theoremstyle{definition}
\usepackage{multirow}

\theoremstyle{remark}

\newif\ifcom
\newtheorem{theorem}{Theorem}

\newtheorem{proposition}{Proposition}
\usepackage{subcaption}
\usepackage{graphicx}
\usepackage{comment}

\begin{document}

\title{Catch Me if You Can: Effective Honeypot Placement in Dynamic AD Attack Graphs}
\author{\IEEEauthorblockN{Huy Q. Ngo}
\IEEEauthorblockA{\textit{Computer and Mathematical Sciences} \\
\textit{University of Adelaide}\\
Adelaide, Australia \\
quanghuy.ngo@adelaide.edu.au}
\and
\IEEEauthorblockN{Mingyu Guo}
\IEEEauthorblockA{\textit{Computer and Mathematical Sciences} \\
\textit{University of Adelaide}\\
Adelaide, Australia \\
mingyu.guo@adelaide.edu.au}
\and
\IEEEauthorblockN{Hung Nguyen}
\IEEEauthorblockA{\textit{Computer and Mathematical Sciences} \\
\textit{University of Adelaide}\\
Adelaide, Australia \\
hung.nguyen@adelaide.edu.au}
}

\maketitle

\begin{abstract}
We study a Stackelberg game between an attacker and a defender on large Active Directory (AD) attack graphs where the defender employs a set of honeypots to stop the attacker from reaching high-value targets. Contrary to existing works that focus on small and static attack graphs, AD graphs typically contain hundreds of thousands of nodes and edges and constantly change over time. We consider two types of attackers: a simple attacker who cannot observe honeypots and a competent attacker who can. To jointly solve the game, we propose a mixed-integer programming (MIP) formulation. We observed that the optimal blocking plan for static graphs performs poorly in dynamic graphs. To solve the dynamic graph problem, we re-design the mixed-integer programming formulation by combining m MIP (dyMIP(m)) instances to produce a near-optimal blocking plan. Furthermore, to handle a large number of dynamic graph instances, we use a clustering algorithm to efficiently find the m-most representative graph instances for a constant m (dyMIP(m)). We prove a lower bound on the optimal blocking strategy for dynamic graphs and show that our dyMIP(m) algorithms produce close to optimal results for a range of AD graphs under realistic conditions.
\end{abstract}


\section{Introduction}

% \hn{The paper reads like a lot of standard algorithms to solve a practically important problem. It won't be enough. We have to explain the cleverness/inventive nature of our solutions. Emphasize on the fact that MIP and all the k-mean, clustering algorithms are major new ideas, ideally suited for this type of problem.}

Microsoft Active Directories (AD) are popular directory services for identity and access management and are deployed at most enterprises. Due to their popularity, AD systems have been major targets for attackers over the last decade. In 2021, Microsoft reported more than 25.6 billion brute force attacks on their AD accounts~\cite{MSRC}. In these attacks, the attacker first builds an attack graph of the targeted AD system where nodes are user accounts, computers, security groups, etc. Each edge in the AD attack graph represents an existing exploit that the attacker can use to move from node to node. The attacker then uses the attack graph to escalate themselves from low privilege nodes to higher privilege nodes (ex. Account A $\xrightarrow[]{\text{AdminTo}}$ Computer B $\xrightarrow[]{\text{HasSession}}$ Account C)~\cite{dunagan2009heat}. Tools for generating AD attack graphs are widely available, such as BloodHound \footnote{https://github.com/BloodHoundAD/BloodHound}. The same tools are also used by blue teams for defending AD systems. Defending AD systems, however, remains very challenging as AD systems are large, complex, and continuously evolve over time. 

In this paper, we study a new method for defending AD by using active defense with honeypots. In our model, the defender places a set of honeypots on nodes of the AD attack graph. The attack campaign fails when the attacker goes into one of the honeypots. We assume that attackers use tools such as SharpHound\footnote{https://github.com/BloodHoundAD/SharpHound} to build the AD attack graph and use the shortest paths from a given set of entry nodes to traverse to the Domain Admin (DA) node. Shortest-path attack plans are provided by default in BloodHound and are widely used by attackers in real attacks~\cite{MITRE2}. 

% mg: maybe add something like if an attacker uses bloodhound as the attacking tool, then by default the attacker will use the shortest path

Active defense with honeypots is not new. However, the problem of placing honeypots in an AD attack graph represents two unique challenges that have not been studied thus far. The first challenge is the scale of the graphs. An AD attack graph typically consists of thousands of nodes and hundreds of thousands of edges with millions of attack paths, even for a small/medium organization. The second challenge is to develop a decoy solution that remains effective even when the graph dynamically changes. AD attack graphs are very dynamic. One of the major sources of changes in the AD graphs is users' activities such as logging on and off a workstation. In an AD attack graph, these dynamics are represented by a special type of edges called HasSession edges \cite{BloodHound}. HasSession edges are added to the graph when a user signs on to a computer and has their credential stored in the computer memory. HasSession edges will stay online until being removed from the graph when the user signs off from the computer after a period of time.

Our solution is designed to be effective against attackers with different tactics and techniques. In our proposed model, we consider two types of attackers: a \textbf{simple attacker}, who is unable to detect any honeypots, and a \textbf{competent attacker}, who employs sophisticated detection tools to obtain complete visibility of all honeypots. In practice, it is difficult to ascertain the level of competence of an attacker. However, we can assume the probability of our network being attacked by one of these attackers based on the observation from previous attacks and/or analysis of the attacker's tactics using databases such as MITRE ATT\&CK~\cite{MITRE}.

%So, we introduced a general defending problem against these two types of attacker agents jointly and denoted as $\mbox{HATK}(\varphi)$, where $\varphi$ ($0\leq \varphi \leq 1$) represents the fraction of competent attacker agents ($1-\varphi$ is the probability of simple attacker agents). 
%We introduce a defending problem denoted as $HATK(\varphi)$, where $\varphi$ is the probability of the network being under attack by the competency attacker. By setting $\varphi = 1$, we focus on defending against pure competent attackers, while $\varphi = 0$ represents defending against pure simple attackers. 

Our first proposal is to use a mixed-integer programming (MIP) method to solve the honeypot allocation problem for mixed-type attackers. For our problem, we observed that the Linear Programming approach can be very efficient in solving AD graphs due to the exploitability of MIP on the unique tree-like structure of AD graphs\cite{guo2022practical}. 

The above solution, however, assumes a static graph and performs poorly when the graph dynamically changes. Our approach to addressing the graph dynamic problem is to simulate all possible variations of the HasSession edges and design a placement solution that works best for all of these graph instances. This solution would only work for very small AD systems with a small number of HasSession edges as the number of graph instances grows exponentially with the number of these edges. Instead, we expand the MIP formulation to include $m$ samples of the AD graph. We call it mixed-integer programming formulation for dynamic graphs (dyMIP(m)). As the run-time of dyMIP(m) grows for $m \gg 1$, we propose 2 heuristics (\textit{voting-based heuristic} and \textit{clustering-based heuristic} that allow dyMIP(m) to approximate the optimal strategy very quickly in these cases. While \textit{voting-based heuristic} is a very natural heuristic based on the idea of running smaller batches of samples on dyMIP(m), hence reducing the $m$ value. \textit{Clustering-based heuristic} approach hypothesizes that there is a number of good-quality samples in our graph sample that could be used to improve the solution for our dyMIP(m) algorithm. The novel idea of \textit{clustering-based heuristic}) is to solve combinatorial in each individual (static) graph to derive features and then apply a clustering algorithm to improve the quality of the sample for the general problem on a dynamic graph. 

% Our first proposal is to use a mixed-integer programming (MIP) method to solve the honeypot allocation problem for mixed-type attackers. The above solution, however, assumes a static graph and performs poorly when the graph dynamically changes. Our approach to addressing the graph dynamic problem is to simulate all possible variations of the HasSession edges and design a placement solution that works best for all of these graph instances. This solution would only work for very small AD systems with a small number of HasSession edges as the number of graph instances grows exponentially with the number of these edges. Instead, we expand the MIP formulation to include $m$ samples of the AD graph. We call it mixed-integer programming formulation for dynamic graphs (dyMIP(m)). As the run-time of dyMIP(m) grows for $m \gg 1$, we propose 2 heuristics (\textit{voting-based heuristic} and \textit{clustering-based heuristic}) that allow dyMIP(m) to approximate the optimal strategy very quickly in these cases.

%To simulate the dynamic of the attack graph, a naive approach involves assuming a random process where the existence of a HasSession edge at any time instance follows a Bernoulli random process. We treat  each HasSession edge as an independently and identically distributed (iid) random variable with fixed probability $p$ of appearance. However, we will demonstrate that this simulation approach for HasSession edge is far from realistic. Instead, our experiment will focus to apply our theoretical model to a more close-to-real-world settings.

Our key contributions are
\begin{itemize}
\item We formulate a general honeypot placement problem in AD graphs considering 2 types of attackers. We show that the static graph version of our problem is already NP-hard when attackers are unable to detect any honeypots and $W[1]$-hard when attackers have complete visibility of all honeypots.
% \item We proposed a studies on the problem of jointly defense against these two types of attackers using honeypot;
\item We provide a mixed integer program formulation and solutions for placing honeypots in very large AD graphs. Our solution could solve the problem on graphs of sizes that none of the previous approaches could; 
% \hn{Can you say something along the line of AD graphs are tree-like, hence MIP works very well? cite Mingyu paper that talks about this?}
\item We study for the first time the honeypot placement problem with dynamic graphs. All of the existing approaches focused only on static graphs. We develop two heuristics based on voting and clustering to solve the dynamic graph problems.  Our solutions, especially the clustering-based method, which derives defense based on ``representative'' snapshots of the dynamic graph (cluster centroids), run very fast on large graphs and provide close to optimal results. Our experiment will consider the algorithm on a range of AD graphs under realistic conditions.

\end{itemize}

\section{Background and Related work}

%There has been some existing work on the AD attack graph. 
Guo et al. \cite{guo2022practical} study the shortest path edge interdiction problem on AD attack graphs. The authors propose a dynamic programming-based tree decomposition method to derive an optimal solution for their edge-blocking problem in directed acyclic AD graphs. They then developed a Graph Neural Network as a heuristic to enhance the scalability of their solution. Guo et al. \cite{guo2022scalable} extend their previous edge-blocking problem by proposing several scalable algorithms including Reinforcement Learning for tree decomposition and mixed-integer programming (MIP) algorithms. The authors later extend the tree decomposition algorithm to a general AD graph with cycles and derive the mixed strategy solution for their problem via MIP. In addition, Zhang et al. \cite{zhang2023oracle} showed the scalability of a double oracle-based algorithm on the same problem. Goel et al. \cite{goel2022defending} assumed the detection and failure probability of attackers on every edge of the AD graph. They deploy a neural network to approximate the attacker's strategy and apply an evolutionary diversity optimization (EDO) strategy to solve the defender problem. In \cite{goel2023evolving}, Goel et al. further improved their defense algorithm (EDO) by improving attacker strategy (fitness function) via Reinforcement Learning. These works, however, only work on static graphs and are not applicable to real AD networks, which are naturally dynamic. One of our key observations in this paper is that optimal strategies on a static graph perform poorly when applied to dynamic networks. Our main difference with these existing works on AD graphs is that our algorithms are designed for dynamic graphs and hence generate better defenses for real AD systems.

% Moreover, while these existing work on AD graph focus on edge removal as defensive tactic, we introduce a honeypot as a new type of defense tactic to this area of research. 

%\hn{Your list of previous works seems short. Is there any paper on honeypot placement in the previous Infocom?}

Honeypots have been widely studied in the literature. Stackelberg games are typically used to formulate the honeypot allocation problem in previous works~\cite{durkota2015optimal,durkota2015approximate,durkota2019hardening,milani2020harnessing}. The models in these studies are very similar to ours: they use attack graphs to model the attacker's actions and the defender can stop the attacker by deploying honeypots in the network. The defender’s task is to minimize the damage caused by the attacker with the minimum cost of honeypot deployment. In \cite{milani2020harnessing}, Milani et al. study games where the defender can either deploy honey-edge or remove an edge from a network to lure the attacker to honeypots. Although these studies can optimally place honeypots in the network, they only work on very small graphs and suffer scalability issues. Indeed, in all of these works, they only consider networks of a hundred nodes in their experiments. Real AD graphs are several orders of magnitude larger than these graphs. Our solutions in this paper are the first solutions that can effectively handle honeypot placement at that scale. 

More recently, Shinde et al. \cite{shinde2021cyber} model the cyber deception game as an interactive partially observable Markov decision process.  The authors of~\cite{shinde2021cyber} focus on luring the attacker to fake data by allocating deception files on network hosts - that is when the attacker already gets in the honeypot. Our work focuses on finding the optimal honeypot allocation strategy at the network level. In \cite{lukas2021deep}, Lukas et al. study the honeypot placement in AD graphs. The author proposed to use deep learning variational autoencoder model to generate and place honey users in the network. Their focus is on the topological structure of the fake AD graphs, not the impact of their solution on the attacker's success rate - a real measure of effectiveness for these solutions. Again, these recent solutions only consider honeypot allocation in small and static networks.

\textit{In this paper, we aim to provide practical solutions for allocating honeypots on large and dynamic AD attack graphs against realistic attackers.}

\begin{figure}[htbp]
\centering
  \includegraphics[width=0.7\linewidth]{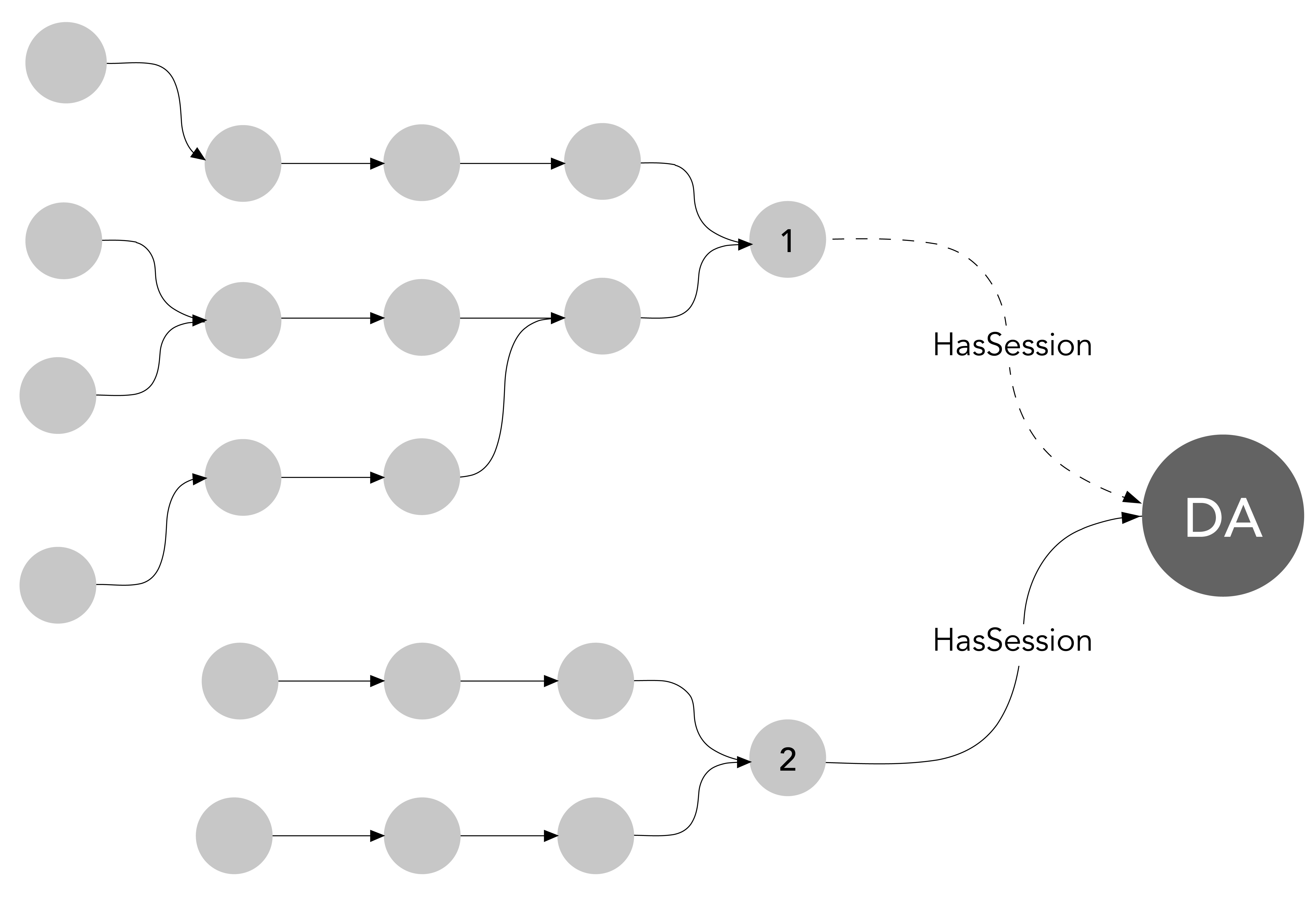}
  \caption{An attack graph with dynamic HasSession edges.}
 \label{fig:dygraph}
\end{figure}

Fig. \ref{fig:dygraph} shows an attack graph of a typical dynamic AD environment. There are two HasSession edges to the DA (The DA uses the edge 1 $\rightarrow$ DA regularly but only logs into 2 when needed). A static snapshot of the graph typically captures only the branch that has 1 $\rightarrow$ DA and completely misses the branch of the graph that contains 2 $\rightarrow$ DA. An attacker would wait in a system for a long time to wait for edge 2 $\rightarrow$ DA to appear and can bypass static defenses built on only the 1 $\rightarrow$ DA branch. None of the previous works on AD attack graphs ~\cite{dunagan2009heat, goel2023evolving, guo2022practical,guo2022scalable,goel2022defending, zhang2023oracle} considered dynamic AD graphs and would fail to defend against this type of attacks.

%\hn{Any previous work on attacker models? We need to stress this point of having more realistic attackers.}

Our solution can provide near-optimal honeypot allocation in dynamic networks with hundreds of thousands of nodes and millions of edges. % \hn{repeat Mingyu's comment here: only a few thousands? I though we could run it on hundreds of thousands or even millions.}
% mingyu: only a few thousands?

% Our paper focus on  can provide near-optimal honeypot allocation in dynamic networks with several thousands of nodes.  

\section{Problem and Model Description}
\label{sec:model}
\subsection{Honeypot placement model for static graphs}
We consider a Stackelberg game between an attacker and a defender on a directed AD attack graph $G = (V, E)$. There is a set $S\subseteq V$ of entry nodes, and the attacker can enter the graph via one of $s = |S|$ entry nodes. The attacker's ultimate goal is to reach the Domain Admin (DA). The attacker performs attacks by choosing paths with pure actions. (i.e.,  an attacker can run BloodHound \cite{BloodHound} to generate the paths to DA). From an entry node, there are many possible attack paths to DA, and the attacker can choose to attack the DA. Also, there is a fixed set of \textit{blockable} nodes, denoted by $N_b \subseteq V$, the defender's task is to pick $b$ nodes in $N_b$ to allocate honeypots in order to intercept as many of the attacker's \textit{attack paths} as possible. A honeypot will stop the attacker from going further into the network, that is, if the attacker stumbles into a honeypot, the attack campaign fails. We study two types of attacker models, the \textbf{simple attacker}  and \textbf{competent attacker}.  We introduce a defending problem denoted as $HATK(\varphi)$, where $\varphi$ ($0\leq \varphi \leq 1$) and represents the fraction of competent attacker agents ($1-\varphi$) is the probability of simple attacker agents).  When $\varphi = 1$, we focus on defending against pure competent attackers, whereas $\varphi = 0$ represents defending against pure simple attackers.

In the simple attacker model, the attacker is assumed to be unable to differentiate between a normal node and a honeypot. The simple attacker will perform the shortest attack path. Therefore, the attacker's optimal strategy is to randomly select one of the shortest paths to the DA. The defender's goal is to reduce the number of ``clean paths'', which are paths that can be used by the attacker to reach the target without any intervention. The defender achieves this by applying a blocking plan $B$ that reduces the number of clean paths. Let's denote by $y_{i}$ the total number of clean paths from entry node $i$ to DA and $y^B_{i}$ as the remaining clean paths after the defender applies a blocking plan $B$. The attacker's success probability is $\frac{y^B_{i}}{y_i}$, and the expected success probability can be obtained by averaging over all entry nodes. Our optimization problem for allocating honeypot against the simple attacker is formally defined as:
$\min\limits_{B \subset N_b, |B| \leq b} \sum\limits_{i=1}^{s} \frac{y^B_{i}}{y_i}$

In the competent attacker model, the attacker is capable of detecting every honeypot in the attack graph. Thus, when the attacker encounters a honeypot on its way to the target destination, it alternates to other paths. To prevent a competent attacker from reaching the DA, the defender must place honeypots to completely disconnect an entry node from the DA. The defender's goal is to maximize the number of entry nodes that are disconnected from the DA using a limited budget. Let's denote $R_{i}$ as the reachability of node $i$ to DA and given that allocating honeypot can change the reachability, we define $R^{B}_{i}$ as the reachability of node i after applying blocking plan $B$. Our optimization problem for allocating honeypot against the competent attacker is formally defined as: $\min\limits_{B \subset N_b, |B| \leq b} \sum\limits_{i=1}^{s} R^{B}_{i}$. Combining both of the attacker model, the problem $HATK(\varphi$) can be interpreted as: 
\begin{equation}
\min\limits_{B \subset N_b, |B| \leq b} \sum\limits_{i=1}^{s} (\varphi R^{B}_{i} + (1-\varphi) \frac{y^B_{i}}{y_i})
\end{equation}
To simplify notation, we introduce the following denotation: $z_{\varphi,i} = \varphi R^{B}{i} + (1-\varphi) \frac{y^B{i}}{y_i}$, where $z_{\varphi,i}$ represents the normalized attacker success rate of reaching the Domain Admin (DA) from node $i$ given a specific $\varphi$ value.

The two theorems below establish the computational hardness of the honeypot placement problems presented in this paper. \emph{Due to space constraints and anonymity, we omit the proofs of these two theorems in this submission. Detailed proofs will be provided in the extended technical report, supplementing the main manuscript.}

\begin{theorem} Let $L$ be the maximum shortest path length from any entry node to DA. 
The static version of the optimal honeypot placement problem against \textbf{Simple Attacker} is NP-hard when $L\ge 7$.
\end{theorem}

\begin{theorem} The static version of the optimal honeypot placement problem against \textbf{Competent Attacker} is $W[1]$-hard with respect to budget b.
\end{theorem}

%\begin{proof} For the detailed proof, please referred to the technical appendix.
%\end{proof}

\subsection{Honeypot placement model for dynamic graphs}

The above formulation only optimizes the honeypot placement for a given graph instance $G$. In real AD networks, the graph $G$ changes constantly due to users' activities. In this paper, we consider dynamic graphs with on/off HasSession edges - a major source of dynamism in AD graphs. We model the dynamic graph process as follows. 

There is a subset of edges $H \subseteq E$ (the HasSession edges) where each edge is turned on and off randomly.  Formally, let $Comp$ be the set of computers and $User$ be the set of users in the network. The authentication data is represented in the tuple format $<t_s, t_e, u_i, c_j>$, where $t_s$ is the sign-in (start time), $t_e$ is the sign-off time (end time), and $u_i$ and $c_j$ are a user and a computer in $User$ and $Comp$, respectively. This tuple indicates that user $u_i$ signs in to computer $c_j$ at time $t_s$ and signs off that computer at time $t_e$. The log-on and log-off time of a session are readily available in typical Windows logs and can also be captured using tools such as SharpHound by attackers. A Hassession between a computer $c_j$ and a user $u_i$ appears in the attack graph $g_t$ at time $t$ ($(c_j, u_i) \in E_{t}$) if $t_s \leq t \leq t_e$.

At time $t$, a snapshot/realization of the dynamic graph is denoted by $g_t = (V,E_t)$, where $E_t = <e_{0,t}, e_{1,t}, \dots, e_{m,t}>$ with $e_{i,t}$ referring to the state of a unique Hassession edge $e_1$ at time $t$. Specifically, $e_{i,t} = 1$ if the edge $e_i$ is online at time $t$, otherwise $e_{i,t} = 0$.

%We assume using reconnaissance tools such as Bloodhound to capture network snapshots at regular intervals $\Delta t$. We remark that a Hassession vulnerability edge capture the following information: computer $c_j$ has the session of user $u_i$. So the directed edge $(c_j, u_i)$ say to be appear in the attack graph $G$ at time $i$ ($(c_j, u_i) \in E_{t}$) if $t_s \leq t \leq t_e$.

%We assume that each HasSession edge is on/off independently of others with a fixed probability. 

%can be obtained by simulating whether each edge in $H$ is on or off. 

Note that only the set of edges $E_t$ changes in $g_t$. The set of nodes $V$ remains the same. We denote the set of all graph instances as $G_s = \{g_1, g_2, g_3, ..., g_m\}$, where $m$ is the number of possible instances. In the dynamic setting, given a defensive budget of $b$, the defender's problem is to allocate honeypots to limit the attacker's clean paths on every possible snapshot/realization of the attack graphs. The dynamic version of the HATK($\varphi$) problem can be interpreted:   
\begin{equation}
\min\limits_{B \subset N_b, |B| \leq b} \sum\limits_{g \in {G_s}} \sum\limits_{i=1}^{s}  (\varphi R^{B}_{i,g} + (1-\varphi) \frac{y^B_{i,g}}{y_{i,g}})
\end{equation}

Again, to simplify the notation, we have: $z_{\varphi,i,g} = \varphi R^{B}{i,g} + (1-\varphi) \frac{y^B{i,g}}{y_{i,g}}$. Note that we do not make any assumption on the temporal pattern of the edge sets in our solutions. We will show in the evaluation section that our algorithms are robust to dynamic graphs with realistic patterns.

% mingyu Typo above, don't think this is the correct formula

% Let $N_{user}$ be the set of user account nodes and $N_{comp}$ be the set of computers in the AD network. There will be $|N_{user}|*|N_{comp}|$ possible HasSession edges in the network which consequently produces a combination of $|G_s| = 2^{|N_{user}|*|N_{comp}|}$ possible graph instances. In real networks, a user usually logs in to their own workstation or some of the regular computers in the network. This makes the number of HasSession edges from computer nodes small. To better model real AD networks, we assume that each user has $log_{10}(|N_{user}|)$ number of HasSession edges on average. This makes the total number or HasSession edges available on each graph follow the binomial distribution $X \sim B(n,p)$ where $n = |N_{user}| \cdot |N_{computer}|$ and $p = \frac{\log_{10} |N_{user}| }{|N_{computer}|}$, $p$ is the probability of a given HasSession edge being online.  

\section{Honeypot Placement Solutions}

\subsection{Solution for static AD graphs}

\subsubsection{Mixed-integer programming for static AD graphs}

Firstly, we formulate a nonlinear program for solving the $HATK(\varphi$):

\begin{subequations}
\begin{align}
\text{min} \displaystyle\sum\limits_{i=1}^{s} z_{\varphi,i}  \nonumber\\ && \nonumber\\
% \end{flalign*}
% \begin{align}
  y^{B}_i =& \sum_{j \in n(i)}y^{B}_{j}, & \forall i \in V \backslash N_b \label{eq:1a} \\
  y^{B}_i =& \sum_{j \in n(i)} (1-B_i) y^{B}_{j}, & \forall i \in  N_b \label{eq:1b} \\
  y^B_{i} \leq& y_{i},         & \forall i \in V \label{eq:1c} \\
  y^B_{i} \geq& 0,              & \forall i \in V \label{eq:1d} \\
  R^B_{i} \geq& R^B_{j} - B_{i}, &(i,j) \in E, i \in N_{b} \label{eq:1e}\\
  R^B_{i} \geq& R^B_{j}        , & (i,j) \in E, \nonumber\\
  &&i \in N \backslash N_{b} \label{eq:1f}\\
  \sum_{i \in V}\ B_{i} \leq& b \label{eq:1g}\\
  B_{i},R_{i} \in & \{ 0,1 \} \label{eq:1h}\\
  R_{DA}, y_{DA} =& 1 \label{eq:1i}
\end{align}
\end{subequations}

Where $n(i)$ is the set of the successor node of node $i$. The design of the formulation is based on the problem of defending against each individual attacker. We formulate it as a nonlinear program that is tailored to each type of agent. In our formulation, we use $B_i$ to denote the unit of budget spent on node i, $B_i$ is binary. Formally, $B_i = 0$ if $i \not \in N_b$ and $B_i \in \{0,1\}$ if $i \in N_b$.

% mingyu: equations numbering seems to be buggy on my end -- the numbers are wrong in pdf, but in source code it seems correct
Constraints (\ref{eq:1a}) to (\ref{eq:1d}) present the problem of defending against the simple attacker. The design of this set of constraints concentrates on computing the number of paths from an arbitrary node to the target (DA) on an \textbf{all-shortest path} graph by summing the number of paths to DA from each of its successors. We illustrate this idea using the example in Fig. \ref{fig:exadgraph}. To calculate the number of paths to DA from 5 or $y_5$, we need to recursively calculate $y_3$, $y_2$, $y_1$ and $y_{DA}$. First, we define that $y_{DA} = 1$ (there is 1 path from DA to itself). Then each node 1 and 2 has exactly 1 direct edge to DA so $y_1 = 1$ and $y_2 = 1$. Next, node 3 has two outgoing edges to nodes 2 and 1, meaning that attackers have two possible ways to reach DA via these nodes. Therefore, $y_3 = y_1 + y_2 = 1 + 1 = 2$. Finally, since node 5 has one outgoing edge to node 3, which could lead to two possible paths to DA, we set $y_5 = y_3 = 2$. Note that our path-calculating approach will fail in the presence of cycles in the graph, as the number of paths of a node in a cycle can go to infinity. Since the simple attacker is unable to observe any honeypot and they always go for the shortest paths, it is without loss of generality to consider only \textbf{all-shortest path} graphs, which is naturally acyclic.
% mg: said something on all shortest path graph
\begin{figure}[h]
\centering
  \includegraphics[width=0.7\linewidth]{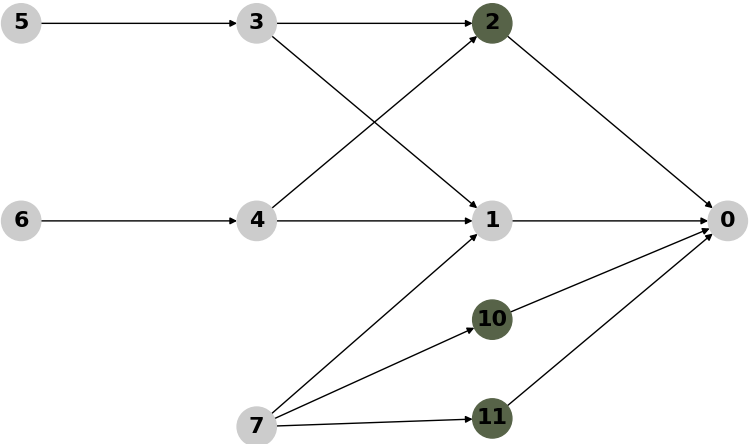}
  \caption{An example all-shortest path AD attack graph. Nodes 5, 6, 7 are entry nodes. Nodes 0 is the DA. \textbf{Bold} nodes 2, 10, 11 are not blockable nodes}
  \label{fig:exadgraph}
\end{figure}

In the case of defending against a competent attacker, we aim to reduce the attacker's success rate by disconnecting the maximum number of entry nodes from DA. Constraints (\ref{eq:1e}) and (\ref{eq:1f}) presented the problem of defending against a competent attacker. These constraints solve the problem based on the idea that if any of the successor nodes of node $i$ can reach DA, then DA is also reachable via node $i$. We assume that DA is reachable by all entry nodes by default.

Our formulation is nonlinear due to the nonlinear constraint (\ref{eq:1b}). To obtain a linear version, we follow several transformation steps. First, we will replace $y_i$ by an auxiliary variable $v_i$ by adding constraints $y^B_{i} \leq  v_{i}$, (\ref{eq:1a}) also become $v_i = \textstyle\sum_{j \in n(i)}y^{B}_{j}$. Next, we replace (\ref{eq:1b}) by 2 additional constraints, $y^B_{i} \leq y_{i}\cdot(1-B_i)$ and $y^B_{i} \geq v_{i} + y_{i}\cdot(-B_i)$. These constraints imply that if node $i$ is blocked, then $y^B_{i} = 0$; otherwise, we compute $y^B_i$ normally ($y^B_i = v_i = \textstyle\sum_{j \in n(i)}y^{B}_{j}$). 

%Due to the space constraints, we will include the complete mixed integer programming in the technical appendix.

\subsection{Honeypot Placement for dynamic networks}
% In this section, we will discuss our proposed graph sampling technique that generalise our algorithm for dynamic graph. We first refining our mixed-integer programming formulation for multiple graph problem. 

\subsubsection{Multi-graph Mixed-integer Programming}
\label{sec:dymip}

% In the previous section, we formulated the mixed-integer programming for the honeypot allocation problem for static graphs. As shown later, blocking strategy for static graph will shortfall in dynamic graph scenario where HasSession edges are changing their states.
For dynamic graphs, we require a different approach as we need to design a blocking plan that performs well against an exponential number of possible graph instances. Assume that we have a collection of ``sample static instances'' (snapshots) based on the dynamic AD graph, denoted by $G_s$. As shown in section \ref{sec:model}, we model the node blocking problem as a global optimization problem where the defender tries to minimize the attacker's average success probability over the sample instances in $G_s$.
%, i.e, defender action of blocking nodes $B$ will imply among difference graph instance in $G_s$. 
Let us denote by $m$ the number of graph samples in $G_s$. A naive approach is to merge all the graphs in $G_s$ (i.e, find a super graph that contains all instances, essentially assuming all HasSession edges are on) and run sMIP on it. Of course, the resulting blocking plan will not be optimal. 

%Another approach to solve the 

%We present our solution for minimizing the attacker's average success probability over the sample instances in $G_s$. In our presentation, we minimize the attacker's sum of success probabilities instead, which is equivalent. 
Let $OBJ(g_i)$ be the objective function on graph $g_i \in G_s$ or the attacker's success probability facing graph $g_i \in G_s$ in the static scenario. 
We also define $OBJ_m(G_s)$ to be the sum of $m$ objective functions or the sum of the attacker's success probabilities for all sample graphs in $G_s$ (there are $m=|G_s|$ samples). 
Optimizing blocking strategy for dynamic graphs then boils down to minimizing the following objective function:

\begin{eqnarray}\label{eq:6}
OBJ_m(G_s) = OBJ(g_1) + OBJ(g_2) + ... + OBJ(g_m)
\end{eqnarray}
In practice, some snapshots of the graph will have a higher chance of appearing, in that case, we could adjust our objective function by multiplying by a weight  $k$  to that graph's objective function (i.e., $k  * OBJ(g_i)$). 
%We could repurpose our earlier MIP for static graphs to handle the task of jointly optimizing for $|m|$ sample graphs.
%Since the set of nodes stays the same, our variable set ($B_i$) stays the same.
%All we need is to merge the constraints on individual graph instances (i.e., simply stack $|m|$ sets of constraints all into the model), and then replace the objective by the sum over individual objectives. 
We have the following nonlinear program for the dynamic version:

\begin{subequations}
\begin{align}
% \text{min} \displaystyle\sum\limits_{g \in {G_s}} \displaystyle\sum\limits_{i=1}^{s} \varphi &R^{B}_{i,g} + (1-\varphi) \frac{y^B_{i,g}}{y_{i,g}}&  \nonumber\\  \nonumber\\
\text{min} \displaystyle\sum\limits_{g \in {G_s}} \displaystyle\sum\limits_{i=1}^{s} z_{\varphi,i,g}&  \nonumber\\  \nonumber\\
% \end{flalign*}
% \begin{align}
  y^{B}_{i,g} =& \sum_{j \in n(i)}y^{B}_{j,g}, & \forall i \in V_g \backslash N_b, g \in G_s \label{eq:3a} \\
  y^{B}_{i,g} = \sum_{j \in n(i)} &(1-B_i) y^{B}_{j,g}, & \forall i \in  N_b, g \in G_s \label{eq:3b} \\
  y^B_{i,g} \leq& y_{i,g},         & \forall i \in V_g, g \in G_s \label{eq:3c} \\
  y^B_{i,g} \geq& 0,              & \forall i \in V_g, g \in G_s \label{eq:3d} \\
  R^B_{i,g} \geq& R^B_{j,g} - B_{i}, &\forall (i,j) \in E_g,  \nonumber \\ 
                &                    &i \in N_{b}, g \in G_s \label{eq:3e}\\
  R^B_{i,g} \geq& R^B_{j,g}        , & \forall (i,j) \in E_g,g \in G_s\nonumber \\
                &                    & i \in N \backslash N_{b} \label{eq:3f}\\
  \sum_{i \in V}\ B_{i} \leq& b \label{eq:3g}\\
  B_{i},R_{i,g} \in & \{ 0,1 \} \label{eq:3h}\\
  R_{DA}, y_{DA} =& 1 \label{eq:3i}
\end{align}
\end{subequations}

The above formulation can be converted to a linear program, similar to the approach involving static graphs. %Due to space constraints, we will only include the complete mixed integer programming in the technical appendix.
In the rest of this paper, we will refer to the Mixed Integer Programming algorithm for dynamic graphs as $MIP_m(X)$, where $X$ represents the set of graph instances considered in MIP and $m$ denotes the cardinality of $X$. We use the term $dyMIP(m)$ when we do not need to specify the set of graphs under consideration.

\subsubsection{Optimal Lower Bound}
% Ideally, to obtained the optimal blocking policy of Sample 
In theory, we can find the optimal blocking policy by running dyMIP(m) algorithm on every possible graph instance. However, the linear program $MIP_m(X)$ requires $\mathcal{O}(m \cdot (n+e))$ variables, where $n$ is the number of nodes and $e$ is the number of edges in the graph $G \in X$. As a result, $MIP_m(X)$ becomes increasingly computationally challenging as $m$ grows larger. To overcome this issue, we split $G_s$ into $j$ equally-sized batches, each batch has $t$ graphs (i.e, $t \cdot j = |G_s|$). We denote each batch of graphs by $X_i \subset G_s$, and $X = \{X_1, X_2, ..., X_j\}$ is the (all disjoint) batch set, where $G_s = X_1 \cup X_2 \cup \cdots \cup X_j$. The following proposition shows the derivation of a Lower Bound on the overall objective using this batching approach.

\begin{proposition} The batching solution gives a lower bound on the overall objective:
\begin{eqnarray}\label{eq:7}
\displaystyle\sum\limits_{X_i \in X} \min_B OBJ_t(X_i) \leq \min_B OBJ_m(G_s) = OPT
\end{eqnarray}
\end{proposition}
%\begin{proof}
% See section 6 in the appendix for the full proof.
%\end{proof}
Again, we omit the proof for space constraints.

For (\ref{eq:7}), the left-hand side consists of sum of the attacker's success probabilities for all batches, and the defender can
apply different optimal blocking plans for different batches. The right-hand side is the sum of the attacker's success probabilities for all samples, under the defender's optimal blocking plan that is applied to all samples (i.e., the same plan shared over all batches).
The inequality in (\ref{eq:7}) provides a lower bound on the optimal result for the set of samples $G_s$. This lower bound can be estimated by summing the optimal results of each batch in $X$.
Even for large values of $m$, this approach allows us to find the estimated lower bound with minimal computational effort. In simpler terms, solving a single instance of $MIP_{1000}(G_s)$ is more challenging than solving 100 instances of $MIP_{10}(X_i)$. By summing the objective values of all instances of $MIP_{10}(X_i)$, we are guaranteed to get a lower bound for the original problem. 
Furthermore, we could interpret the result of each batch as {\em an independent draw from the same (unknown) distribution} and our task is to calculate the mean based on sample averages,
so using Hoeffding’s Inequality Bound (more details presented below), we can achieve a high-confidence estimation without going over all batches.

%In our case, the objective value represented the chance of the attacker can get to the DA. It is noteworthy that the objective function is the sum of the attacker's success probability in every graph. For the representation, we will show the normalised result which takes the average over the number of possible graphs. 

% The above program has at most $O()$

% \textbf{Voting for best blocking strategy:} 

\subsubsection{Generalization to Large Collections of Graph Samples}

As mentioned above, dyMIP(m) only works on a small number of graphs before running into scalability obstacles. We need to develop a better strategy for scaling dyMIP(m) to large $m$. We develop two heuristics for this purpose.

\vspace{0.75cm}
%\subsubsection{Voting based heuristic}
\noindent \textbf{{Voting based heuristic}}

The overall idea behind our \textit{voting based heuristic} is to make use of the lower bound calculation in Section \ref{sec:dymip}. First, we divide $G_s$ into $j$ equal disjoint batches and run $MIP_t(X_i)$ on each batch and run each $MIP_t(X_i)$ to derive a blocking strategy $B_i$. Denote by $B = \{B_1, B_2, \cdots, B_j \}$ (the set of blocking strategies generated based on all $MIP_t(X_i)$ instances). We use $C = (c(1), c(2), \cdots, c(n))$ to count the number of times an individual node $n$ is blocked (selected as a honeypot) under the blocking plans in $B$. We then sort $C$ in descending order and pick the top $b$ nodes in $C$ as the final blocking/honeypot placement strategy. In other words, we solve for $j$ separate blocking plans
based on $j$ batches, and use ``majority voting'' to come up with a blocking plan (find the $b$ most voted nodes to place honeypots).

\noindent \textbf{Clustering based heuristic}

Our second heuristic is based on the idea that there is a set of most \textit{representative} graph instances (snapshots, realizations) for the whole collection of samples. This approach is particularly useful when there are duplicate graphs in a set of samples or samples with similar structures. Our idea is only to run dyMIP(m) once on this \textit{representative} set of graphs. To do that, we apply a clustering algorithm to explore different structures among the graph samples and group graphs with common structures into the same group. Formally, Let's define $feature(g)$ as features of graph $g$: $feature(g) = \big( f(1,g), f(2,g), \cdots, f(s,g) \big )$, where $f(i,g) = \frac{y^{B_g}_{i,g}}{y_{i,g}}$ represents the attacker's success rate when they are at a starting node $i$ and facing a node blocking strategy $B_g$. $B_g$ is produced by applying staticMIP on graph $g$. We use k-means as the clustering algorithm in our implementation. After getting $feature(g)$ for every graph $g \in G_s$, we apply k-means to cluster $G_s$ into $k$ clusters. We consider graphs that are closest to centroids as the representative graphs for our dyMIP(m), where the distances to centroid are Euclidean distances~\cite{danielsson1980euclidean}. We apply dyMIP(m) on the set of representative graphs that we got from the clustering algorithm. 
% \begin{eqnarray}\label{eq:7}
% feature(g) = \bigg( \frac{y^{B_g}_{1,g}}{y_{1,g}}, \frac{y^{B_g}_{2,g}}{y_{2,g}}, \cdots, \frac{y^{B_g}_{i,g}}{y_{i,g}} \bigg )
% \end{eqnarray}

% \begin{eqnarray}\label{eq:7}

% \end{eqnarray}

There is a complication with the clustering algorithm where clusters are of varying sizes and densities. If we uniformly pick a graph from each cluster, then the algorithm is prone to be biased to the ``fringe'' clusters which degrade the quality of our samples. Our solution is to weigh the graph selection based on the cluster size. That is, when choosing a graph for dyMIP(m), we first choose $m$ clusters where each cluster $c$ with size $size_c$ has a probability $\frac{size_c}{|G_s|}$ for being picked. We then pick a graph instance that has the closest distance to centroid $c$ to be the \textit{representative graph}. The already chosen graph instances are removed from consideration to avoid duplication. \textit{Clustering heuristic} is more scalable than \textit{voting heuristic}. For example, suppose our batch size is $100$, while \textit{voting heuristic}  requires us to run dyMIP(100) multiple times (majority voting does not make sense if there is only one vote), the \textit{clustering heuristic} only requires us to run MIP(100) once.

\section{Evaluation}
We evaluate our proposed solutions on both synthetic AD graphs with both randomized HasSession edges and real HasSession data. All of the experiments are carried out on a high-performance computing cluster with 2 CPUs and 24GB of RAM allocated to each task for the experiments. Code and data will be made publicly available.

\subsection{Synthetic Graph Generation}
%Since the Active Directory attack graph data is confidential, we assess the performance of our algorithms using synthetic data. 
We start with synthetic graphs generated by DBCreator \footnote{https://github.com/BloodHoundAD/BloodHound-Tools/tree/master/DBCreator} and Adsimulator \footnote{https://github.com/nicolas-carolo/adsimulator} - two state of the art tools for creating AD graphs.
In particular, we present results for the 5 synthetic graphs in Table \ref{tab:compgreedy}). The graphs labeled as "Rxxxx" were generated using DBCreator, while the ones labeled as "ADxxx" were created using Adsimulator. DBCreator allows us to generate graphs given a number of computers in the targetted AD network (R2000 and R4000 have 2000 and 4000 computer nodes, respectively). Adsimulator provides more flexibility by allowing graphs with specific properties, including the number of users, computers, domain trusts, OUs, GPOs, and more. $ADX05$ and $ADX10$ are created by increasing all its default parameters by a factor of 5 and 10 respectively. $ADU10$, $ADU100$ are created by tuning the parameters to mimic the real AD network of {\em University of Anonymized} which are $10\%$, $100\%$ of the size of the mimicked network, respectively.

We perform the following pre-processing steps, First, when there are multiple DAs in the graph, we simply merge it into 1 unique DA. We only consider 3 types of edges in our graph: AdminTo, HasSession, and MemberOf. We only consider an entry node that can reach DA. $ADU100$ is the largest graph in our experiment with 137207 nodes and 1410679 edges. Table \ref{tab:graphstat} show the size of each graphs
%We referred the reader to the appendix for details of the size of each graph in our experiment. 

\subsection{Experiments on static graphs}
\subsubsection{Experiment Set Up}
We will refer mixed-integer programming approach as sMIP (or staticMIP). 
We also adopt 2 algorithms referred to as sMIP-S, sMIP-C where the former is sMIP formulation for solving the problem of (pure) simple attacker agent ($\varphi = 0$), and the latter is sMIP for solving the problem of (pure) competent attacker agent ($\varphi = 1)$. 
Throughout our experiment, we assume $\varphi = 0.5$ (equal chance of being attacked by either of the attackers). However, it is important to note that our technique remains effective for any value of $\varphi$. 
%In summary, our experiment will have three settings for $\varphi$ (0, 0.5, 1), which will be denoted as sMIP-S, sMIP-M, and sMIP-C, respectively.

% We also have another setting for dealing with mixed types of attackers and the MIP algorithm for solving that problem will be denoted as sMIP-M. We argue that the competent attacker, in the most extreme case, have a higher capability to infiltrate our network compared to the simple attacker. In our experiment, we assume a ratio of $s:1$ (or $\varphi =\frac{s}{s+1}$) between the competent attacker and the simple attacker. This signifies that the more the entry node, the greater the chance the network is vulnerable with competent attackers. 

 We use Gurobi 9.0.2 solver for solving staticMIP. For each graph, we run 10 trials and report the results for these runs. For each trial, we randomly draw 50 entry nodes and set the budget b to 10. We assumed only computer nodes are blockable (i.e., honeypots can be placed only on computers). In the experiment, we introduce 3 metrics to evaluate our algorithms: \textbf{S}imple-Agent \textbf{S}uccess \textbf{R}ate (\textbf{SSR}) and \textbf{C}ompetent-Agent \textbf{S}uccess \textbf{R}ate (\textbf{CSR}) and \textbf{M}ean \textbf{S}uccess \textbf{R}ate (MSR). SSR and CSR are the success rates of the Simple and Competent attacker agent when facing a blocking strategy from the defender. SSR is measured as the fraction of the clean paths over the number of paths in the all-shortest path graph, normalized by the number of starting nodes and CSR is the fraction of entry nodes that can reach DA over the total number of entry nodes; whereas MSR is the average of SSR and CSR. 
 %Basically, SSR is the fraction of the clean paths over the number of paths in the all-shortest path graph, normalised by the number of starting nodes, CSR is the fraction of entry nodes that can reach DA over the total number of entry nodes; and MSR is mean of SSR and CSR. 

We carry out 2 sets of experiments. The first experiment which shows in Table \ref{tab:compgreedy} compares our sMIP-S and sMIP-C with 2 greedy algorithms namely GREEDY-S, GREEDY-C. 
Additionally, we evaluated our approach against Zhang et al.'s Double Oracle \cite{zhang2023oracle} (ZDO) algorithm. ZDO is the state-of-the-art edge-blocking algorithm for AD graphs. We adopt this approach for honeypot placement by formulating it as a simple blocking problem.  The purpose of this comparison is to demonstrate that the existing techniques are not suitable for our honeypot placement problem without compromising its performance.

%which is a completely different problem to ours. We re-implemented their node-blocking version and compared it to ours. However, it is still noteworthy that our paper is not simply the node blocking version of the previous works, we study a completely new defense technique that is honeypot placement problem which is completely difference from other works (difference objective function). The purpose of this comparison is to demonstrate that the existing techniques are not suitable for our defense approach without compromising its performance.

In the GREEDY-S algorithm, for each node, we can enumerate the shortest path that goes through it or calculate the betweenness centrality for each node. The idea is to iteratively block the node that has the highest betweenness centrality and repeat it $b$ times. GREEDY-C uses a minimum node cut algorithm to disconnect each entry node and then prioritizes the entry nodes requiring the least budget to be blocked until the budget is depleted. In the second experiment shown in Table \ref{tab:mixedstatic}, we simply show the performance of our sMIP against a scenario of 2 types of attacker agents in the network.

\subsubsection{Result Interpretation} In Table \ref{tab:compgreedy}, it can be seen that sMIP-S and sMIP-C perform at least as well as, and often better than, GREEDYs on various graphs. While ZDO performed badly since it is designed for the shortest path interdiction problem. In Table \ref{tab:mixedstatic}, our goal is to produce a defence plan that can be effective against both types of attacker agents. 
As they do not make any relaxation in the MIP formulation, both sMIP-S and sMIP-C obtained the optimal defense against a Simple attacker and a Competent attacker, respectively.
As shown sMIP-M significantly improves the MSR for $\varphi = 0.5$. This is not surprising as it is specifically designed to defend against both types of attackers. In $ADU100$, 10 honeypots are not enough to stop the competent attacker from any entry nodes hence, the attacker success rate for the competent attacker is 1. More significantly, staticMIP runs fast enough for large graphs and could scale to graphs of sizes equivalent to our anonymous network. That is, staticMIP is able to produce optimal results for very large static graphs, making it scalable for realistic AD graphs. Note that even though the general problem is NP-hard, what we observe here is that AD graphs have certain tree-like structures \cite{guo2022practical} that allow staticMIP to solve them very fast. In fact, the defending problem against both types of attackers is polynomially solvable if the AD graph is a tree (via a top-down Dynamic Program from DA to allocate honeypots). This makes the Linear Programming approach very efficient against problems on tree-like AD graphs.

% \hn{tree-like, cite Mingyu paper that talks about this?}\hn{Explain why? what in the property of the problem that allows staticMIP to solve an NP-hard problem very fast?}. 

We note, however, that staticMIP is not designed to work with dynamic graphs --- a topic that we will address next.

\begin{table}
\begin{center}
\caption{Comparing sMIP algorithm against GREEDY algorithms. SSR is for simple attackers and CSR metric is used for competent attackers. Here $b=10$.}\label{tab:compgreedy}
\smallskip\noindent
\resizebox{\linewidth}{!}{%
\begin{tabular}{|l||ccc||cc|}
\hline
       & \multicolumn{3}{c||}{\textbf{Simple Attacker (SSR)}}       & \multicolumn{2}{c|}{\textbf{Competent Attacker (CSR)}}                   \\
\cline{2-6}
                  & sMIP-S                 & GREEDY-S           & ZDO      & sMIP-C                  & GREEDY-C       \\
\hline
$\textbf{R2000 }$ & \textbf{0.233} (0.15s)          & \textbf{0.233}  (0.01s)     & 0.389 (0.67s)      & \textbf{0.354} (0.09s) & 0.542 (0.37s) \\
$\textbf{R4000 }$ & \textbf{0.279} (0.10s)          & \textbf{0.279 } (0.01s)     & 1.000 (0.13s)      & \textbf{0.530} (0.28s) & 0.560 (0.48s) \\
$\textbf{ADX05 }$ & \textbf{0.135} (0.12s) & 0.136  (0.02s)     & 0.272 (0.68s)      & \textbf{0.670} (0.12s) & 0.682 (1.04s) \\
$\textbf{ADX10 }$ & \textbf{0.144} (0.16s) & 0.155  (0.03s)     & 0.226 (0.48s)      & \textbf{0.606} (0.20s) & 0.774 (1.86s) \\
$\textbf{ADU10 }$ & \textbf{0.547} (1.27s)          & \textbf{0.547}  (0.27s)     & 0.717 (0.54s)      & 0.984 (1.29s)          & 0.984 (24.5s) \\
$\textbf{ADU100}$ & \textbf{0.602} (15.3s)          & \textbf{0.602}  (3.80s)     & 0.760  (0.15s)      & 1     (-)              & 1    (-) \\
\hline
\end{tabular}}
\end{center}
\end{table}

\begin{table}
\begin{center}
\caption{Comparison between variants of sMIP algorithm. Optimal results are in \textbf{bold}.}\label{tab:mixedstatic}
\smallskip\noindent
\resizebox{\linewidth}{!}{%
\begin{tabular}{|l||ccc||ccc||ccc|}

\hline
       &\multicolumn{3}{c||}{sMIP-S}&\multicolumn{3}{c||}{sMIP-M}&\multicolumn{3}{c|}{sMIP-C}            \\
\cline{2-10}
       & \textbf{SSR} & \textbf{CSR} & \textbf{MSR} & \textbf{SSR} & \textbf{CSR} & \textbf{MSR} & \textbf{SSR} & \textbf{CSR} & \textbf{MSR}   \\
\hline
$\textbf{R2000}$  & \textbf{0.233}   & 0.682 &  0.46  & 0.303   & 0.356   &  \textbf{0.330}  & 0.313   & \textbf{0.354}  & 0.334     \\
$\textbf{R4000}$  & \textbf{0.279 }  & 0.856 &  0.568 & 0.421   & 0.536   &  \textbf{0.478}  & 0.452   & \textbf{0.53}   & 0.498\\
$\textbf{ADX05}$  & \textbf{0.135}   & 0.958 &  0.547 & 0.17    & 0.794   &  \textbf{0.482}  & 0.492   & \textbf{0.67}   & 0.581    \\
$\textbf{ADX10 }$ & \textbf{0.144}   & 0.992 &  0.568 & 0.409   & 0.63    &  \textbf{0.519}  & 0.468   & \textbf{0.606}  & 0.539   \\
$\textbf{ADU10 }$ & \textbf{0.547}   & 1     &  0.774 & 0.547   & 1       &  \textbf{0.774}  & 0.912   & \textbf{0.984}  & 0.953    \\
$\textbf{ADU100}$ & \textbf{0.602}   & 1     &  0.801 & 0.602   & 1       &  \textbf{0.801}  & 0.851   & \textbf{1}      & 0.925  \\
\hline

\end{tabular}}
\end{center}

\end{table}
% mingyu no explanation so far on table 2, I would include the "final" susccess rate, i.e., 0.5*SSR+0.5*CSR in table and make that the most prominent numbers for comparison
\subsection{Experiments on synthetic dynamic graphs}
\label{sec:dynamicex}

% mingyu, change to In table ?, we present if this paragraph is describing the above table
In Table \ref{tab:dynamic}, we present our experiments on the $ADX05$ graph with a budget of 20. Only computer nodes are blockable. We consider all user nodes in the graph as the entry nodes. After preprocessing, the numbers of user nodes and computer nodes are 72 and 500 respectively. Each edge is randomly ON in each run with probability $p = 0.5$. Ideally, we will run our dyMIP(m) on every possible instance to get the optimal blocking policy. However, this is computationally infeasible. A more practical way is to derive a blocking policy on a training set (sampled from the network) and evaluate it on a different testing set (future states of the network). In our experiments, we use 5,000 instances for training and run a Monte Carlo simulation on 100,000 graphs to evaluate the effectiveness of our blocking plan to ensure statistically significant results for the evaluation.

%In the next part, we will explain why 100,000 samples are enough to ensure statistically significant results for the evaluation.
%In our Monte Carlo Simulation, by testing our blocking polity on $n =$ 100,000 graph samples. We proved that the results test on 100,000 can be at least 99\% confident with error within $\epsilon = 5.147\times 10^{-3}$ of the true mean (proof provided in the technical appendix.

\subsubsection{Estimating the Optimal Lower Bound}
We compute the Lower Bound in~\eqref{eq:7} for the AD graphs using a sample size of 100,000 graphs. 
%Based on the Central Limit Theorem, we can be confident in treating this result as an accurate estimate of the Lower Bound. Each batch in this experiment can be viewed as an independent draw, and the Lower Bound is calculated as the average over multiple batches.
We find a Lower Bound of CSR and SSR on a graph using dyMIP-$C(m)$  and dyMIP-$S(m)$  respectively. We set $m=50$ and run it 200 times to cover all 100,000 graph instances in Monte Carlo Simulation.
\textbf{After conducting this experiment on ADX05, the estimated the Lower Bound for SSR is  $0.7240\pm0.0055$ and for CSR is $0.8889\pm 0.0019$.}
These estimates are used to measure the optimality of our solutions.

\subsubsection{Discussion of Dynamic Results}

\begin{table}
\begin{center}
\caption{Experiments on dynamic graphs ADS05. We show the success rate of each algorithm with respect to their intended attacker (e.x. We show the Simple attacker Success Rate (SSR) for dyMIP-S algorithm, etc.)}\label{tab:dynamic}
\smallskip\noindent
\resizebox{\linewidth}{!}{%
\begin{tabular}{|l|l|l|l|l|l|}

\hline
      % \multicolumn{9}{c||}\textbf{dyMIP-S} \\
    & &\multicolumn{1}{c|}{m = 1$^{\mathrm{a}}$} & \multicolumn{1}{c|}{m = 10} & \multicolumn{1}{c|}{m = 50}   & \multicolumn{1}{c|}{m = 100}        \\
% \cline{2-7}
  % \textbf{dyMIP-S}     & \textbf{SSR}     & \textbf{SSR}     & \textbf{SSR}  & \textbf{SSR}     \\
\hline
\multirow{3}{*}{dyMIP-S} &\textbf{Rand} & 0.7960 (2s)       & 0.7430 (16s)     & 0.7359 (89s)       & 0.7326 (174s)    \\
                         &\textbf{Kmean}  &                   & 0.7409 (919s)    & 0.7343 (987s)      & 0.7325 (1062s)         \\
                         &\textbf{Vote} & 0.7304 (10232s)   & 0.7293 (8403s)   & 0.7293 (9095s)     & \textbf{0.7293} (9643s)        \\
\hline

\hline

\multirow{3}{*}{dyMIP-M}   &\textbf{Rand} & 0.8851 (2s)        &0.8494 (17s)       & 0.8461 (98s)           & 0.8407 (216s)     \\
                          &\textbf{Kmean}  &                    & 0.8544 (741s)    & 0.8398 (1011s)         & 0.8373 (1310s)   \\
                          &\textbf{Vote} & 0.8417 (2416)      & 0.8391 1716       & 0.8353 2136            & \textbf{0.8353} (2666s)  \\
\hline

% \textbf{dyMIP-C}   & \textbf{SSR} & \textbf{CSR}      & \textbf{SSR} & \textbf{CSR}      & \textbf{SSR} & \textbf{CSR}  & \textbf{SSR} & \textbf{CSR}     \\
\hline
\multirow{3}{*}{dyMIP-C} &\textbf{Rand}    & 0.9722 (2s)    &  0.9028 (16s)       & \textbf{0.8889} (95s)      & \textbf{0.8889} (188s)     \\
                         &\textbf{Kmean}     &                &  0.9000 (949s)      & \textbf{0.8889} (1010s)    & \textbf{0.8889} (1069s)    \\
                         &\textbf{Vote}    & 0.9042 (9188s) &  0.9028 (8718s)     &  \textbf{0.8889} (10684s)  &  \textbf{0.8889} (12878s)    \\
       \hline

\multicolumn{6}{l}{$^{\mathrm{a}}$ Random with m = 1 is staticMIP (static graph's policy on the dynamic graph)}

\end{tabular}}
\end{center}

\end{table}

Table~\ref{tab:dynamic} summarizes our experimental results. Each experiment is conducted with 10 trials. We run our algorithm with 3 settings: pure simple attacker (dyMIP-S and $\varphi = 0$), pure competent attacker (dyMIP-C and $\varphi=1$) and mixed-type attacker (dyMIP-M and $\varphi=0.5$).  We randomly draw a training set of 5,000 graphs for each trial. 
We also run the 2 proposed heuristics: voting-based heuristic (\textit{Voting}) and clustering-based heuristic (\textit{Kmean}). We also include a third algorithm (\textit{Random}), which randomly picks $m$ graphs from the samples and applies MIP(m) to that $m$ graphs only.  
For each setting, we set $m$ to 1, 10, 50, and 100 (note that Kmean does not make sense for $m=1$ since one cluster is not meaningful).
In Table~\ref{tab:dynamic}, note that \textit{Random} with $m = 1$ is equivalent to running the MIP algorithm on one static graph (\textit{staticMIP}), and \textit{Voting} with $m = 1$ is applying the voting heuristic on \textit{staticMIP}.
 
 For \textit{Voting}, we run it with 10 parallel tasks. For example, \textit{Voting} with $m = 50$ with 5,000 graphs will require 100 rounds of MIP(m) if we run it sequentially, which is infeasible without parallel computing. The reported time for Voting is the sum of the time of 10 parallel processes. 
As \textit{Random} and \textit{Kmean} only require running MIP(m) once,  we run these algorithms without parallelization. 
For evaluation, we test our algorithms on 100,000 randomly drawn graph samples. %We also split our testing into 10 parallel tasks to save time. 

\textbf{Result Interpretation:} We observed that static defense policy (staticMIP or Random with $m=1$) in all 3 $\varphi$ values performs poorly against all types of attackers.
We are able to obtain optimal CSR (= 0.8889) for various settings with dyMIP-C and dyMIP-M.
While for SSR (= 0.7293), we are able to obtain near-optimal results with dyMIP-S using Voting heuristic and m = 100.
Our experiments demonstrated that \textit{Kmean} outperforms \textit{Random} in defending against simple attackers (yielding better SSR) under various settings. 
This shows that the clustering method (\textit{Kmean}) can improve the ``quality'' of the samples and therefore improve the performance of MIP(m). 
Notably, when only considering settings that can obtain the optimal policy against a competent attacker (CSR = 0.8889), Kmean with m = 100 outperform others in term of SSR (= 0.7781).  
% mingyu I don't understand the above sentence
As the voting heuristic requires running dyMIP(m) on every possible graph in the training set to perform majority voting. Therefore, \textit{Voting} easily outperforms the other methods in terms of SSR as it scales.
However, this makes it computationally more expensive compared to Kmean and Random, which only require running dyMIP(m) once. Therefore, Kmean is more favourable when computational resources are limited. We remark that the success rate obtained from this experiment is quite high for the budget of 10. It is due to the high number of HasSession edges makes the graph very dense (each snapshot could contain $\approx 36,000$ HasSession edges) and hard to defend with 10 honeypots. However, in the next experiment, we will show that our approach could obtain a manageable success rate in a realistic attack graph.

\subsection{Experiment on real dynamic AD graphs}

\begin{figure}
\centering
\begin{subfigure}{0.23\textwidth}
    \includegraphics[width=\textwidth]{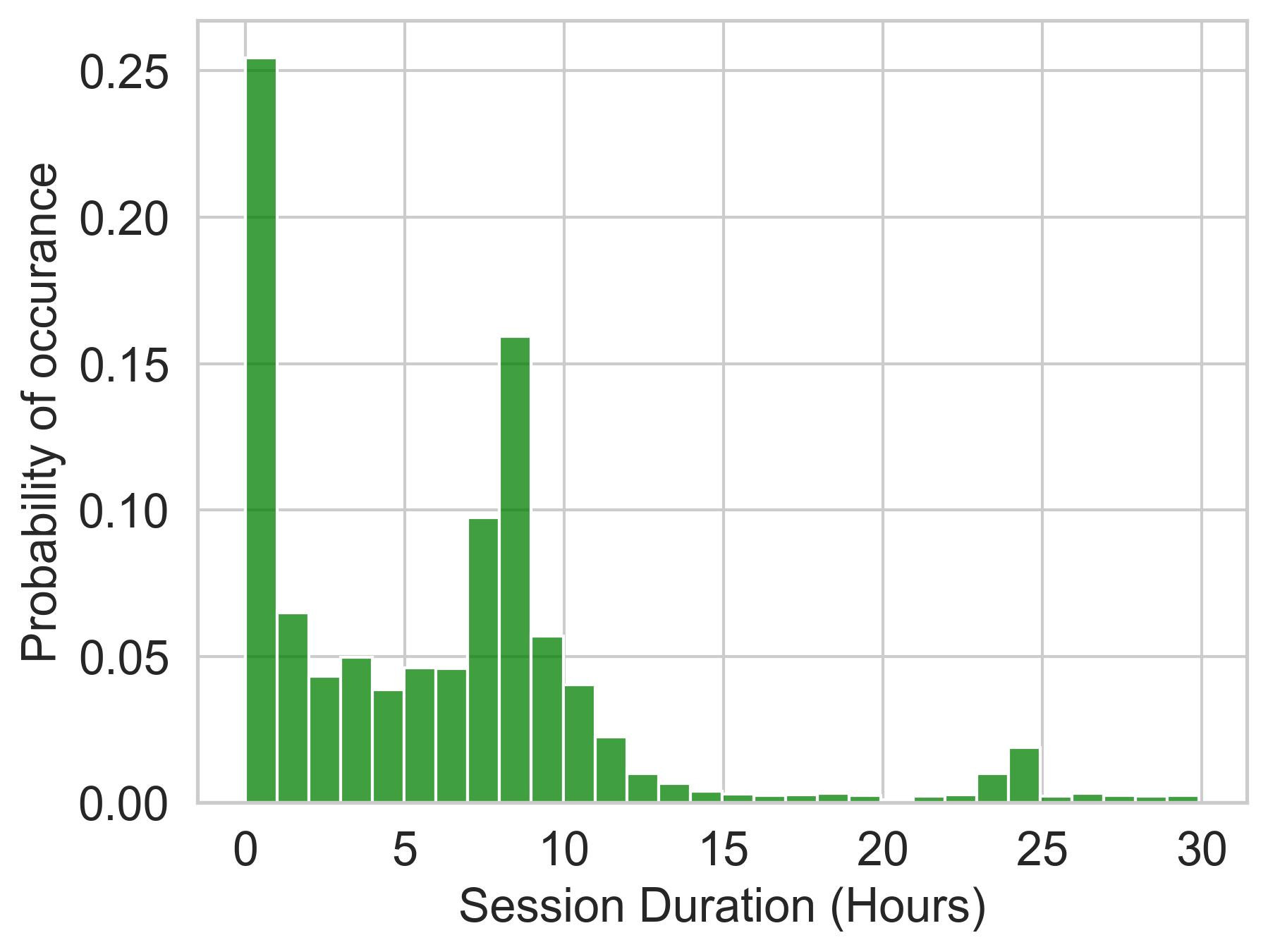}
    \caption{}
    \label{fig:session_length}
\end{subfigure}
\hfill
\begin{subfigure}{0.23\textwidth}
    \includegraphics[width=\textwidth]{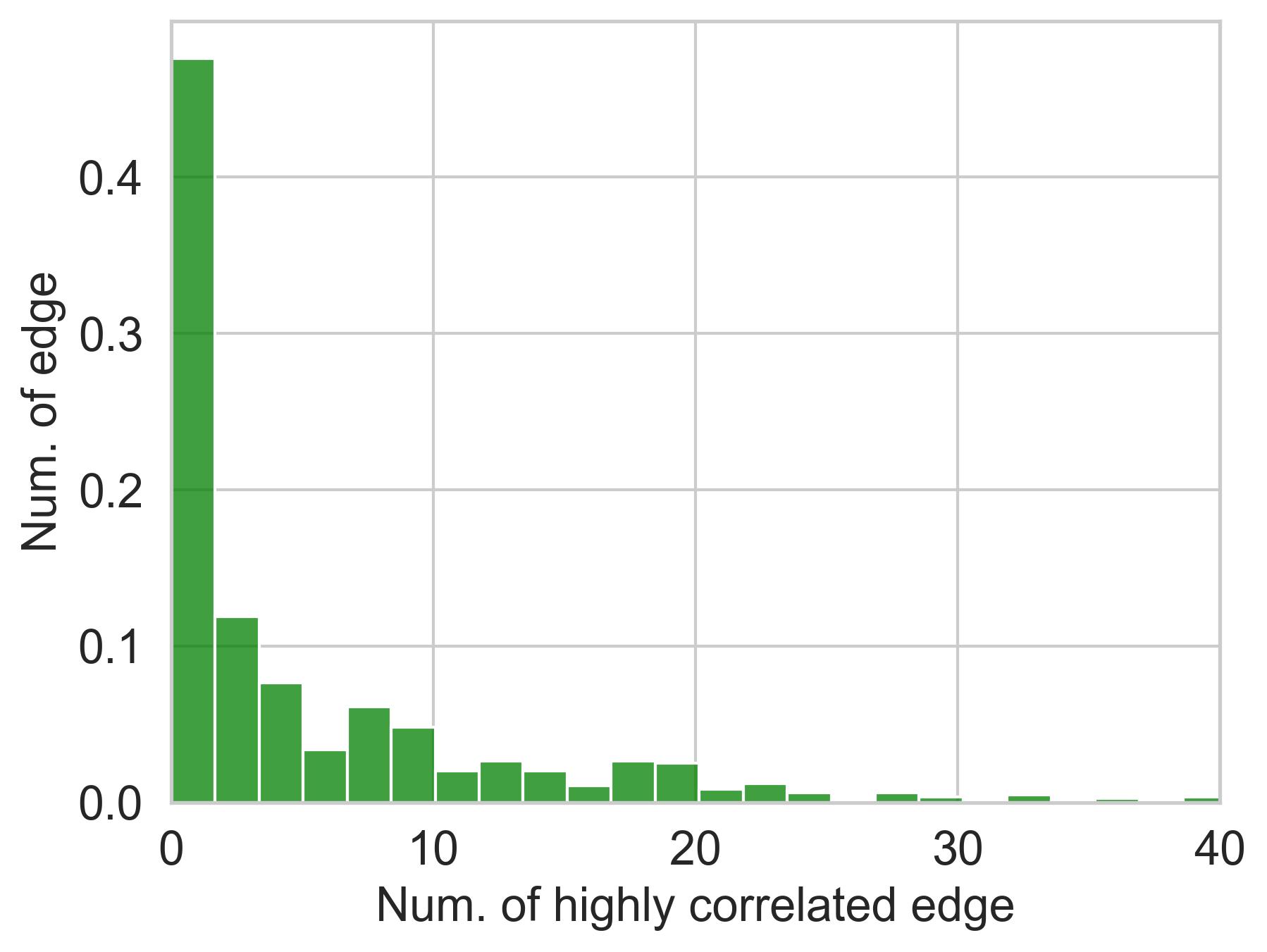}
    \caption{}
    \label{fig:correlation}
\end{subfigure}     

\caption{(a) Number of online sessions and newly established session sampling every 1 hour (b) Histogram of a number of edges which are highly correlated (in terms of logon/off session pattern) with others.}
\label{fig:figures}
\end{figure}

\begin{figure}
\centering
\begin{subfigure}{0.24\textwidth}
    \includegraphics[width=\textwidth]{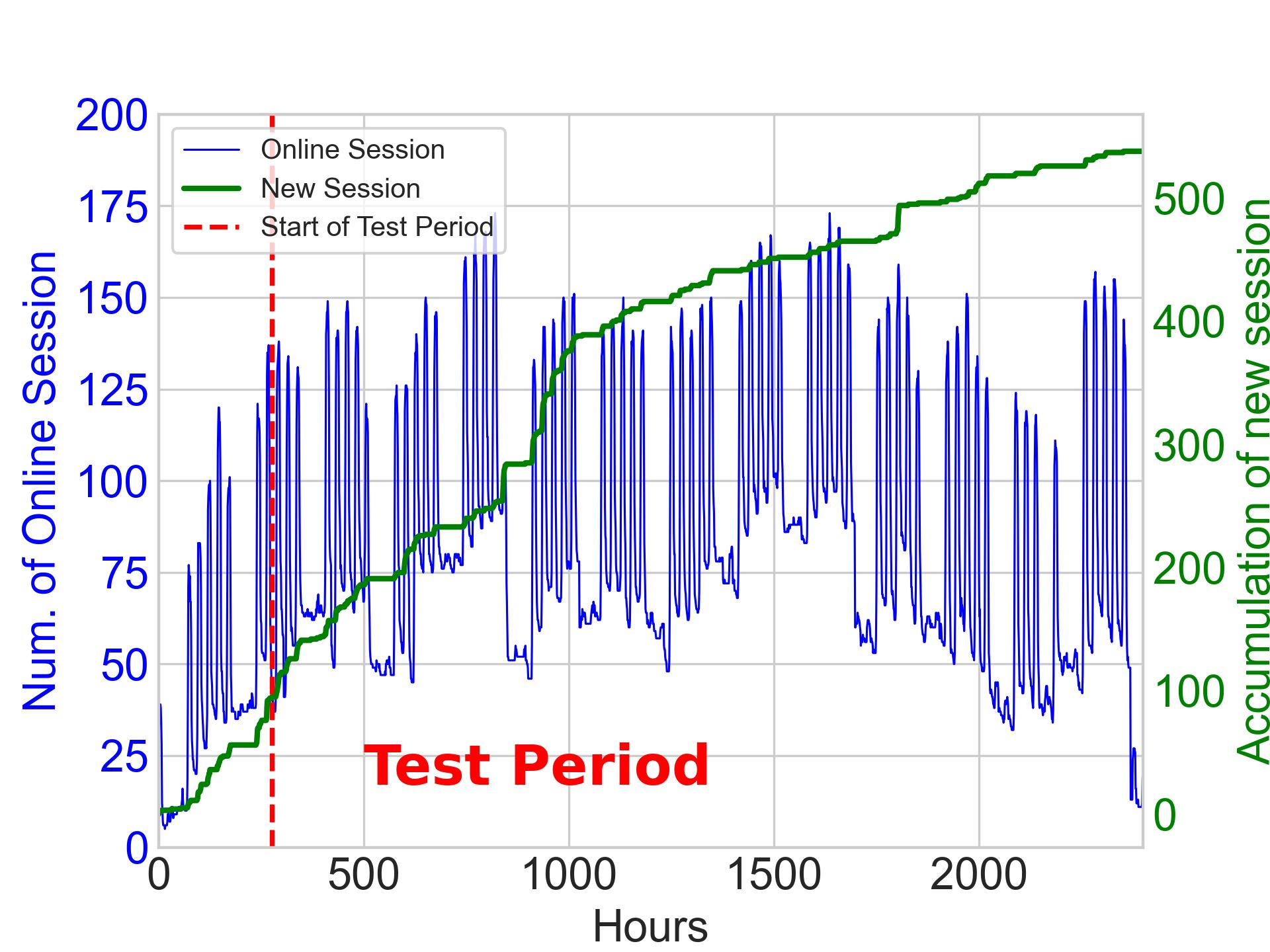}
    \caption{}
    \label{fig:online_session}
\end{subfigure}
\hfill
\begin{subfigure}{0.24\textwidth}
    \includegraphics[width=\textwidth]{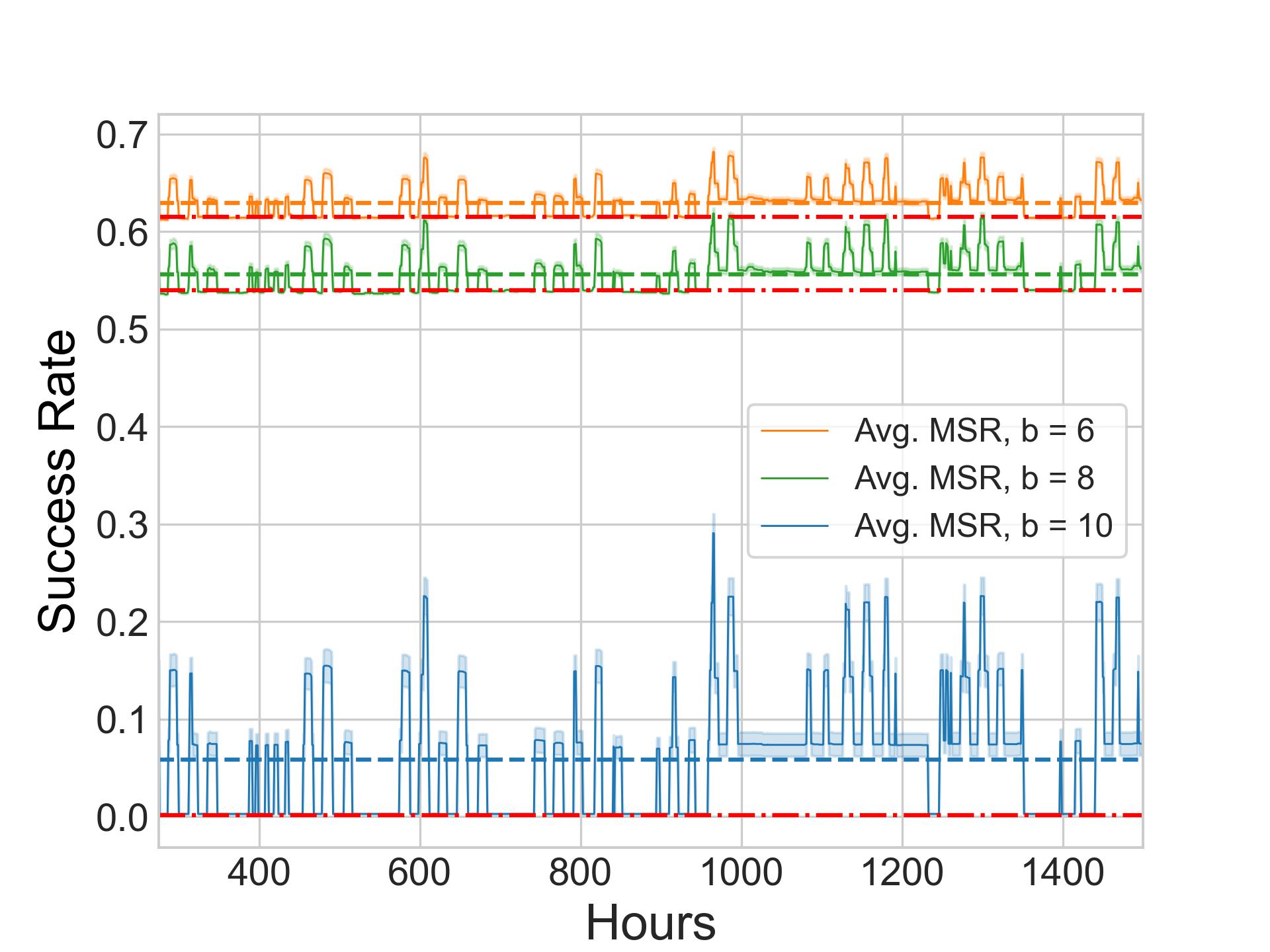}
    \caption{}
    \label{fig:resultcomp}
\end{subfigure}     

\caption{(a)  Number of online sessions and the accumulation of newly established session sampling every 1 hour (b)Voting with m = 10. The red dash-dotted lines are the optimal MSR of b = ${6, 8, 10}$, listed from top to bottom.}
\label{fig:figures}
\end{figure}

\subsubsection{Analyzing authentication data}
%In this section, we design a close-to-real-world experiment using authentication event data. We use the authentication event data to simulate the dynamism of the Hassession edges.
We analyze in this section the temporal properties of HasSession edges in a real network and apply these properties to simulate realistic dynamic AD graphs.

 The dataset used in this section is sourced from an anonymous organization and referred to as \textbf{AuthOrg} data. The dataset includes the authentication events observed over a 102-day period (from 24-06-2022 to 04-10-2022). There are 270 users, 338 computers, 14249 sessions and 1015 unique sessions in the dataset. Upon analyzing the \textbf{AuthOrg} data, we find that the random authentication scheme in Section \ref{sec:dynamicex} significantly diverges from realistic session patterns. Our analysis reveals that the activities of Hassession edges are not independent; rather, they exhibit correlations with each other.

From Fig. \ref{fig:online_session}, the graph exhibits a denser number of Hassession edges during working hours, indicating a higher likelihood of their appearance during this period. 
% The average number of Hassession edges appearing in a snapshot generated using authentication data is still notably sparser (approximately 81 sessions/snapshot) compared to the random authentication scheme in Section \ref{sec:dynamicex} (approximately 45,000 sessions/snapshot). 
Secondly, Hassession edges can persist online for a period of time; for instance, some normal employee's session lasts approximately 8 hours before sign-off (Fig. \ref{fig:session_length}). Thirdly, in Fig. \ref{fig:correlation}, we present a histogram of highly correlated session profiles for each edge. For every edge $e_i$, we count the number of session profiles in the set $E_t\backslash e_i$ that exhibit a high correlation with $profile(e_i)$ 
% \hn{This needs to be rewritten to be precise}.
We define two vectors as \textit{highly correlated} if their Pearson correlation is greater than $0.5$. Fig. \ref{fig:correlation} provides evidence that a significant portion of the edges exhibits high correlations with at least one other edge where the percentage of edge profiles that have at least 1 or more highly correlated edges $P[X \geq 1]$ is more than $63.8 \%$.
%\hn{What is the Pearson Correlation number? Give it here}. %($P[X \geq 1] = 0.638$).

%As shown on our analysis, the random authentication scheme in Section \ref{sec:dynamicex} exhibits significant deviations from a realistic graph when temporal factors of user behaviors are ignored. Consequently, when deploying our algorithm on graph with real authentication data, there are potential factors that may degrade our algorithm's performance. 
%Correlated HasSession edges could degrade the performance of our honeypot placement algorithms in two major ways.
%Firstly, the number of online edge in each snapshot (or the density of edges in the graph) could negatively impact the algorithm's effectiveness. As observed in Fig. \ref{fig:online_session}, there are the increase in the number of session during the work hours resulted from employee come to work. As the graph becomes more populated with Hassession edges, the number of paths to the Domain Admin (DA) increases, subsequently elevating the attacker's success rate.
In addition to correlated edges, another factor that could affect the performance of our honeypot placement algorithm is the number of new or rare link connections. New connections is connection that the establishment of unique sessions for the first time (edges that did not exist before).
The accumulative new connection session line in Fig. \ref{fig:online_session} illustrates a continuous establishment of new sessions throughout the entire dataset duration. These new links can introduce alternate attack paths that were not considered during the training period, thereby contaminating the graph with unexpected paths and degrading the performance of the defense policy over time.
We show in the next section that our algorithms by sampling representative behaviours across time horizons are robust to these degradation factors. 
%\hn{Anything else that helps our algorithms?}

%In this experiment, we will consider these factor to evaluate the effectiveness of our algorithm.

\subsubsection{Simulated AD graphs with realistic HasSession edges} 
We augment the Adsimulator AD graphs in the previous sessions with the log-on and log-off data to generate more realistic AD graphs. Here, we randomly map users and computers from the authentication data to the synthetic graph which we refer to as the "AuthOrg" graph. %While this method overlooks the correlation between user patterns and the overall network topology, such as the difference in login behaviors of IT administrators and lower privilege users, it represents the best viable option for our experiment. 
We train our algorithm using the first 250 hours of data, excluding the initial 25 hours for training stability. The subsequent 2,073 hours are used for testing, resulting in a total of 2,448 hours of data. For ease of comprehension, the figures in this paper only display results up to 1,500 hours. We take snapshots on the graph every 0.25 hours, so we have 1000 snapshots for the training in total. 
%Initially, we attempted blocking only computer nodes, but in the AuthOrg graph, this led to an infeasible solution where the MSR remained 1 even if all computer nodes were blocked. So alternatively, we use another relaxation scheme which assumes that 
In the experiments, 80\% of nodes are blockable (this number could be changed and didn't affect the overall results). We used the voting heuristic with $m = 10$ for all experiments. Additionally, each set is tested on 10 trials.

%\subsection{Degradation Factors.} As shown on our analysis, the random authentication scheme in Section \ref{sec:dynamicex} exhibits significant deviations from a realistic graph when temporal factors of user behaviors are ignored. Consequently, when deploying our algorithm on graph with real authentication data, there are potential factors that may degrade our algorithm's performance. 
%Firstly, the number of online edge in each snapshot (or the density of edges in the graph) could negatively impact the algorithm's effectiveness. As observed in Fig. \ref{fig:online_session}, there are the increase in the number of session during the work hours resulted from employee come to work. As the graph becomes more populated with Hassession edges, the number of paths to the Domain Admin (DA) increases, subsequently elevating the attacker's success rate.
%Another factor is the appearance of new or rare link connections. New connections is connection that the establishment of unique sessions for the first time (edges that did not exist before).
%The new connection session line in Fig. \ref{fig:online_session} illustrates a continuous linear increase in new sessions throughout the entire dataset duration. These new links can introduce alternate attack paths that were not considered during the training period, thereby contaminating the graph with unexpected paths and degrade the performance of the defense policy over time.
%In this experiment, we will consider these factor to evaluate the effectiveness of our algorithm.

\subsubsection{Result Interpretation.} In Fig. \ref{fig:resultcomp}, we present the results for varying the budget, specifically $b = {6, 8, 10}$. Our algorithm demonstrates strong performance with low optimality gaps for each setting, with gaps of 0.0129, 0.0189, and 0.049 for $b = 6, 8,$ and $10$, respectively. We are able to archive the low attacker success rate of 5.83 $\%$ on average with a budget of 10. Notably, we observed that the attacker's success rate is influenced by two main factors: the number of online edges in the network and the presence of certain new link connections (rare links) established in the network.

The experimental results exhibit a high correlation with the number of online sessions plotted in Fig. \ref{fig:online_session}. For instance, with the $b = 10$ setting, the Pearson correlation between the number of online sessions and the success rate over time is 0.5870. %\hn{What does this mean?} 

Our algorithm shows overall resilience against the effects of new link connections when they are introduced linearly over time. The (Pearson) correlation between the number of new connections and the success rate over time, with $b = 10$ setting, is only 0.07. Still, we observed a burst of new connections occurring throughout week 6 (from hour 800th to 1000th, Fig. \ref{fig:online_session}), which had a noticeable impact on the overall algorithm performance, as visually observed in Fig. \ref{fig:resultcomp}. % \hn{Explain what causes this? rare links?} 
Our hypothesis is that during these periods there are \emph{rare new links} that the $m$ sample graphs do not capture. 
% \hn{Why don't you try to increase $m$}

Overall, the algorithm performs well with low optimality gaps. However, it tends to perform worse during periods with an increased number of connections in the network, such as working hours. However, it is explainable that as the graph becomes more populated with Hassession edges, the number of paths to the Domain Admin (DA) increases, subsequently elevating the attacker's success rate. Another reason is that we also assumed that we have a limited number of honeypots and having more honeypots will suppress this trend. Additionally, the presence of rare certain new link connections during the test period can also affect the algorithm's performance. This limitation can be addressed by periodically retraining the algorithm to adapt to changing network conditions or increasing the number of snapshots. 
% \hn{Show results of this!!!}

Finally, Table~\ref{tab:graphstat} shows the scalability of our algorithms.  As shown, the largest network that we applied our algorithms to (ADU100) has 137,315 nodes and 1,4900766 edges. This is several orders of magnitude larger than typical graphs that any existing algorithms for honeypot placement can handle.
\begin{table}[H]

\begin{center}
\caption{This table provides the size of each graph in terms of the number of nodes, edges, user nodes, and computer nodes.}\label{tab:graphstat}
\smallskip\noindent
\resizebox{\linewidth}{!}{%
\begin{tabular}{|l||c||c||c||c|}

\hline
       & \textbf{Nodes}   & \textbf{Edges}    & \textbf{Users } & \textbf{Computers} \\
       \hline
\textbf{R2000 } & 5997   & 18795   & 2000  & 2001 \\
\textbf{R4000}  & 12001  & 45780   & 4000  & 4001 \\
\textbf{ADX05 } & 1624   & 6955    & 482   & 526  \\
\textbf{ADX10}  & 3123   & 15767   & 972   & 1026 \\
\textbf{ADU10}  & 13834  & 94878   & 6372  & 366  \\
\textbf{ADU100} & 137315 & 1490766 & 63172 & 3378 \\
\textbf{AuthOrg}& 2624   & 18594   & 1804  & 326 \\
       \hline
\end{tabular}}
\end{center}
\end{table}

\section{Conclusion}

This paper investigated a Stackelberg game model between an attacker and a defender on Active Directory attack graphs. The model considers two types of attackers, one of which can observe honeypots while the other cannot. The defender aims to allocate honeypots on attack graph nodes to minimize the attacker's expected success chance to reach DA. We show that the problems for both types of attackers are computationally hard and propose a mixed-integer programming (MIP) formulation to solve the problem when there is a mixed-types of attacker agents exist in the network, for both static and dynamic graphs. Our experiments showed that MIP scales well for a graph with $\approx$ 100 thousand nodes and $\approx$ 1 million edges. We show that our algorithm that combines m MIP (dyMIP(m)) instances of the graphs performs well under realistic temporal patterns. Our experimental results show that the MIP(m) algorithms produce near-optimal blocking plans, with the help of two heuristic methods based on majority voting and k-means clustering.

%In practice, AD attack graphs are dynamic as edges can appear or disappear dynamically over time. We showed that defense obtained on static graphs performs poorly on dynamic graphs. We extended our model to consider the dynamic changes of HasSession edges by combining m MIP (dyMIP(m)) instances to produce a better blocking plan. We proved a lower bound on the optimal blocking strategy for dynamic graphs.  

% \begin{appendix}

% \end{document}

\newpage
\newpage

% \begin{comment}

\section{Appendix}
\subsection{Proof of Theorem 1}

%Here in this section, we are going provide the proof for the honeypot allocation problem in AD attack graph. 
% This proof inspired theorem 3.9 of L-length bound node cut in \cite{baier2010length}. L is the length of the shortest path to the destination path to DA
\begin{theorem}
%\textit{\textbf{Theorem 4.1:} 
Let $L$ be the maximum shortest path length from any entry node to DA. 
The static version of the optimal honeypot placement problem against \textbf{Simple Attacker} is NP-hard when $L\ge 7$.
\end{theorem}

% \begin{proof} 

Proof: Let $G_{vc} = (V_{vc}, E_{vc})$ is the vertex cover instance and $G_{vc}$ is undirected graph. Let us define $L$ as the maximum shortest path length from entry nodes to DA node. We construct attack graph as follow. Let $n = |V_{vc}|$. We construct $s$ entry nodes and a single destination DA. For every nodes $i \in V_{vc}$, we construct 6 nodes which is not blockable: $i_1$, $i_2$, $i_4$, $i_5$ and $i_6$. 3 blockable nodes (h-type node): h-type1 ($h_{i1}$), h-type3 ($h_{i3}$) and h-type5 ($h_{i5}$). We called $H$ is set of blocked nodes. We create a set of multiple edge (each with $M$ parallel edges where $M$ is assume to be a large value) which connect each pair of node: ($s_i$, $h_{i1}$), ($h_{i1}$, $i_2$), ($i_2$, $h_{i3}$), ($h_{i3}$, $i_4$), ($i_4$, $h_{i5}$) and ($h_{i5}$, $i_6$); and set of single edge: ($s_i$, $i_1$), ($i_1$, $i_2$), ($i_4$, $i_5$) and ($i_5$, $i_6$). Assume node $i$ and node $j$ connected in $V_{vc}$, we then create two cross path in our gadget from $i_2$ to $j_4$ (via $c_{j3}$) and $j_3$ to $i_4$ (via $c_{i3}$).

Let note that in our gadget, if we choose 1 node to block, for example node $h_{i1}$. We will guarantee that at least $M$ paths being removed from $s_1$ and $i_2$. We refer every path that is not blocked by any honeypot as "clean path". To recall, in our model, to maximise the chance of attack get into the honeypot, we will try to limit number of available "clean path" in the graph by choose number of nodes in the graph to allocate the honeypot. 

Let $V'_{vc} \subseteq V_{vc}$ be a vertex cover and $x = |V'_{vc}|$ is the vertex cover size. Giving the defensive budget $b = n_{vc} + x$, the best way to spend defensive budget is to locate the vertex cover of size $x$. Optimal blocking plan could be obtained by the following \textbf{implication}: If $i \in V'_{vc}$ we add 2 blockable node h-type1 (left) and h-type5 (right) to our blocking set $H$ and for each node $u \in V_{vc} \backslash V'_{vc}$, we add 1 node h-type3 (middle) to blocking set $H$. 

For every $i \in V_{vc}$, there must be at least 1 budget spend on each corresponding path gadget $i$ on gadget graph or there will be at least $(M^2+1)^3$ path to DA via this gadget. Next, if one node of the gadget in $H$ then it must be node of type-2; otherwise it will be at least $(M^2+1)^2$ path via this gadget. Now, if two or more honeynode of gadget in $H$, then it must be type-1 and type-3; this will guarantees that at most $Z$ path available via this gadget. Finally, following the implication, assume a path gadget $ i \in V'_{vc}$ and $d$ is the node degree of $i$ (in $G_{vc}$). 

In this optimal scheme, number of "clean path" is at most $(M^2+1)(2d+1)$. Also, in the optimal scheme, we try to avoid the sub-optimal case of $\mathcal{O}((M^2+1)^2)$ or $\mathcal{O}(M^4)$. Then we have:

\begin{eqnarray}\label{eq:s}
\mathcal{O}(M^4) > \mathcal{O}((M^2+1)(2d+1)) = \mathcal{O}(2dM^2) \nonumber
\end{eqnarray}
\begin{eqnarray}\label{eq:ss}
\Rightarrow \mathcal{O}(M) > \mathcal{O}(\sqrt{2d}) \nonumber
\end{eqnarray}

In this proof, we assume that multi edge exist in our graph, the similar proof for the case of non-multi-edge graph can be obtained by substituting every edge (except edges to DA) with a node and 2 edge to and from these new nodes.

\begin{figure}[h]
  \includegraphics[width=1.1\linewidth]{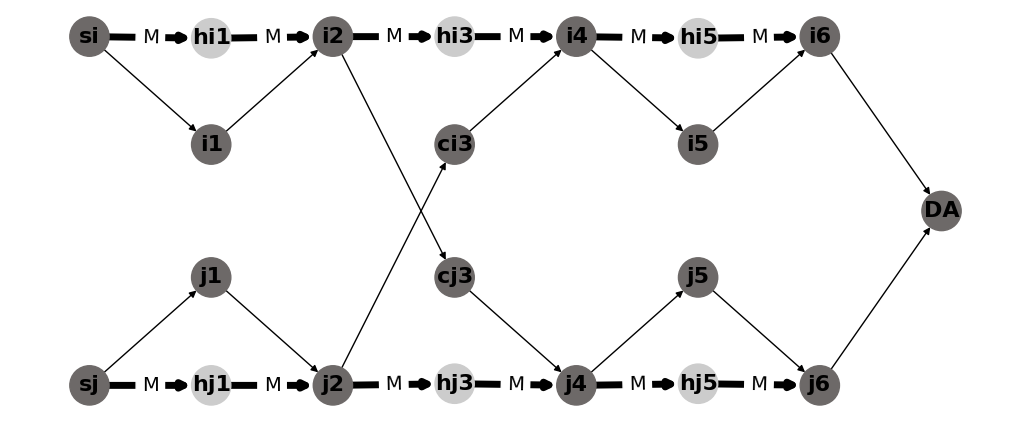}
  \caption{Proof gadget for Theorem 1. Light nodes are blockable nodes. Thick edge is multiple edges ($M$ of them) and thin edge are single edge.}
  \label{fig:NPgadget}
\end{figure}
% \end{proof}

\subsection{Proof of Theorem 2}
\begin{theorem}
The static version of the optimal honeypot placement problem against \textbf{Competent Attacker} is $W[1]$-hard with respect to budget b .
\end{theorem}
% \begin{proof}
Proof: To prove the hardness of our problem, we use the reduction from \textit{clique} problem to our. Clique problem is know to be $W[1]-hard$ with respect to clique size \cite{fellows2009parameterized}. Let define undirected graph $G_{c} = (V_{c}, E_{c})$ is the clique instance. We construct attack graph $G$ as follow. For every edge $(i,j) \in E_{c}$, we construct a corresponding entry node $s_{i,j}$. For every node $v \in G_{c}$ we construct a pair of node $v_{ub}$ and $v_{b}$ in the attack graph instance where $v_{ub}$ is unblockable node. Next, for every entry nodes, we create 2 path $s_{i,j}\rightarrow i_{ub} \rightarrow i_{b} \rightarrow DA$ and  $s_{i,j}\rightarrow j_{ub} \rightarrow j_{b} \rightarrow DA$. If we want to block all path from an entry node $s_{i,j}$ to DA, we need to block both $i_{b}$ and $j_{b}$. Let's say we can choose up to $b$ node to block, the optimal blocking plan could be obtained by locate the clique of size $b$ in $G_{c}$. By allocating a clique, we can disconnect maximum of $b(b-1)/2$ entry nodes.

% \end{proof}
\subsection{Greedy Algorithm for simple attacker agent}

We start with a greedy algorithm to give the reader a better understanding of our model. The greedy algorithm also serves as a baseline model. Assume that  we have an all-shortest paths AD as in Figure \ref{fig:exadgraph}. (An all-shortest path AD graph is a graph where every edge is part of at least one shortest path from an entry node to DA. Edges that are not part of any shortest path can be ignored in our model, so we focus only on all-shortest path AD graphs.) 
% mg: add the definition, please check whether this is what you mean
In this example, we have some open nodes for allocating honeypots (blockable nodes): 1, 3, 4. In this graph, there are 3 entry nodes and 7 open paths to DA: 2 paths from entry node 5, 2 paths from entry node 6 and 3 paths from entry node 7. The attacker's success rate is initially 1 (100 $\%$). Let's say we choose to allocate a honeypot at node 1, i.e, if the attacker chooses any path through node 1, the attack will end and is considered as unsuccessful. There are 2 paths DA via node 1: $5 \rightarrow 3  \rightarrow 1  \rightarrow 0$ and 6 $\rightarrow 4  \rightarrow 1  \rightarrow 0$. Both of these paths have $\frac{1}{3}*\frac{1}{2}=\frac{1}{6}$ probability of being chosen by the attacker to attack the DA. So, by blocking node 1, the defender has $\frac{1}{3}$ chance of blocking the attacker from attacking the DA and the attacker success probability is reduced to only $\frac{2}{3}$.

\begin{figure}[htbp]
\centering
  \includegraphics[width=0.8\linewidth]{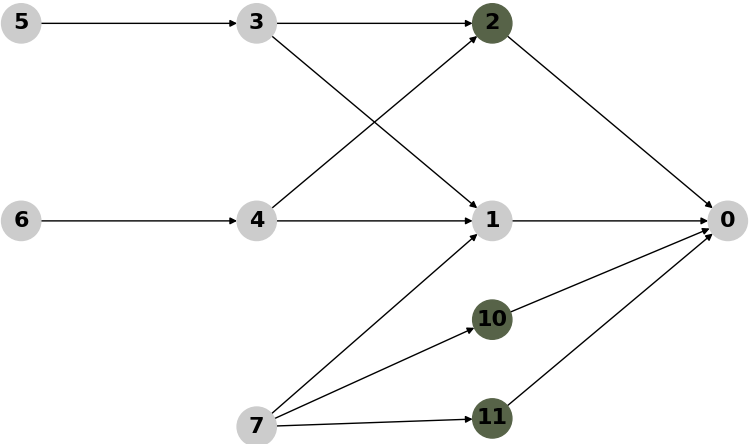}
  \caption{An example all-shortest path AD attack graph. Nodes 5, 6, 7 are entry nodes. Nodes 0 is the DA. Bold nodes 2, 10, 11 are not blockable nodes. We assume DA and entry nodes are naturally not blockable.}
  \label{fig:exadgraph}
\end{figure}

The main idea of the Greedy algorithm is to iteratively block the node that has the largest impact on reducing the expected attacker's success probability and repeat it $b$ times. We observe that the Greedy algorithm performs poorly when there are \textit{substitutable block-worthy} nodes. For example, the graph in Figure \ref{fig:exadgraph} is where Greedy algorithm  produces a sub-optimal result for a blocking budget of $b=2$. On this graph, consider 3 blockable nodes 3, 4, 1. Greedy will pick either the pair of nodes $(1, 3)$ or $(1, 4)$ to block, while the optimal blocking plan is $(3,4)$. Now, Let us run through the example. In the first iteration, node 1 will be blocked as its blocking score is the largest which is $\frac{1}{3}\cdot \frac{1}{3}+\frac{1}{3}\cdot\frac{1}{2}+\frac{1}{3}\cdot\frac{1}{2} = \frac{4}{9}$ compared to $\frac{1}{3}\cdot\frac{1}{2} + \frac{1}{3}\cdot\frac{1}{2} = \frac{1}{3}$ for node 3 and node 4. In the second iteration, node 3 and node 4 are equally block-worthy as their blocking scores are both $\frac{1}{3}\cdot\frac{1}{2} = \frac{1}{6}$. Finally, the blocking plan by Greedy is (1,3) or (1,4), reducing attacker success probability to $1 - \frac{4}{9} - \frac{1}{6} = 0.389$. As mentioned above, the optimal blocking plan on this graph is to block the pair (3,4), giving an attacker success rate of only $1 - \frac{1}{3} - \frac{1}{3} = 0.334$. 
% Figure \ref{fig:exadgraph} is the example of the reduced shortest path attack graph where 5, 6 and 7 are entry nodes. Node 0 is the destination. Light nodes are blockable nodes (bold nodes 2, 10 and 11 are unblockable). 
% Greedy sub-optimal solution: (1,4) or (1,3), Optimal solution: (3,4)
% Enumerate all of the possible shortest path in the graph, this could be done quite efficiency when we only consider shortest path from entry nodes. Next, iterate through all node and update their blocking score, blocking score of node v is determine by summing the chance of all path that include node v. i.e the chance the attacker \textbf{passing} this node. Based on the score with the given budget b, we pick the node with highest blocking score to the reduce the chance attacker get to DA. We repeat it b times.

% \begin{table}[t]
%   \caption{List of path to DA and their path score of Figure topology}
%   \label{tab:locations}
%   \begin{tabular}{lll}\toprule
%     \textit{label} & \textit{path} &  \textit{path score} \\ \midrule
%     $[1]$ & $\langle 5, 3, 2, 0 \rangle$ & $\frac{1}{}$ \\
%     $[2]$ & $\langle 5, 3, 1, 0 \rangle$ & 1580 \\
%     $[3]$ & $\langle 6, 4, 2, 0 \rangle$ & 1765 \\
%     $[4]$ & $\langle 6, 4, 1, 0 \rangle$ & 5 \\
%     $[5]$ & $\langle 7, 1, 0 \rangle$ & 18 \\
%     $[6]$ & $\langle 7, 10, 0 \rangle$ & 18 \\
%     $[7]$ & $\langle 7, 11, 0 \rangle$ & 18 \\ \bottomrule
    
%   \end{tabular}
% \end{table}

\subsection{MIP formulation for finding total of path to DA}
We provide the formulation for calculating the
number of shortest paths from all nodes in the graph without any
blocking plan:

\begin{subequations}
\begin{align}
\text{min}  \displaystyle\sum\limits_{i}^{s} y_i && \nonumber\\
            y_i =& \sum_{j \in n(i)} y_{j}, \ & \forall i \in N \label{fig:eq3a}\\
            y_{DA} =& 1
\end{align}
\end{subequations}

% \subsection{Extension for MIP folmulation}

% Constraint (\ref{eq:4a}) implies that if a node in consideration is blocked, then the node value $y^B_{i}$ will be reset to 0. (\ref{eq:4a}) constraint is non-linear as it involves multiplication of a binary variable $B_i$ and continuous variable $y^B_j$. To linearize the above terms, we introduce auxiliary binary variables $v_i$ and re-write the earlier non-linear formulation as below. Constraints (\ref{eq:5a})-(\ref{eq:5f}) is the linearization for (\ref{eq:4a})

% \begin{subequations}
% \begin{align}
% \text{min}   \displaystyle\sum\limits_{i}^{s} \frac{y^B_{i}}{y_{i} \cdot s}&& \nonumber\\
% y^B_{i} =& (1 - B_i)\sum_{j \in n(i)}y^{B}_{j}, & \forall i \in N \label{eq:4a}\\
%                 y^B_{i} \leq& y_{i},         & \forall i \in N\\
%                 y^B_{i} \geq& 0,              & \forall i \in N\\
%                 \sum_{i \in N_{b}}\ B_{i} \leq& b, & N_b \subset N\\
%                 %  &                                        B_{i} \in \{ 0,1 \} \\
%                 B_{i} \in& \{ 0,1 \} \\
%                 y^B_{DA} =& 1
% \end{align}
% \end{subequations}
\subsection{Complete formulation for Multi-graph MIP}
We define $v_{i,g}$ as the temporary variable for $y_{i,g}$
\begin{subequations}
\begin{align}
\text{min} \displaystyle\sum\limits_{g \in {G_s}}\displaystyle\sum\limits_{i=1}^{s}& \varphi R^{B}_{i,g} + (1-\varphi) \frac{y^B_{i,g}}{y_{i,g}} & \nonumber\\  \nonumber\\
% \end{flalign*}
% \begin{align}
  v_{i,g} =& \sum_{j \in n(i)}y^{B}_{j,g}, & \forall i \in V_g \backslash N_b, g \in G_s \label{eq:1a} \\
  % y_{i,g} =& \sum_{j \in n(i)} (1-B_i) y^{B}_{j,g}, & \forall i \in  N_b, g \in G_s \label{eq:1b} \\
  % y^B_{i,g} \leq& y_{i,g},         & \forall i \in V_g, g \in G_s \label{eq:1c} \\
  % y^B_{i,g} \geq& 0,              & \forall i \in V_g, g \in G_s \label{eq:1d} \\
  y^B_{i,g} \leq& v_{i,g}, & \forall i \in N_b, g \in G_{s} & \\
  y^B_{i,g} \leq& y_{i,g} (1-B_i), & \forall i \in N_b, g \in G_{s}\\ 
  y^B_{i,g} \geq& v_{i,g} + y_{i,g} \cdot (-B_i) ,& \forall i \in N_b, g \in G_{s}\\
  y^B_{i,g} \geq& 0 ,& \forall i \in N_b, g \in G_{s}\\
  y^B_{i,g} =& v_{i,g},& \forall i \in N_b, g \in G_{s}\\
  R^B_{i,g} \geq& R^B_{j,g} - B_{i}, &\forall (i,j) \in E_g,  \nonumber \\ 
                &                    &i \in N_{b}, g \in G_s \label{eq:1e}\\
  R^B_{i,g} \geq& R^B_{j,g}        , & \forall (i,j) \in E_g,\nonumber \\
                &                    & i \in N \backslash N_{b}, g \in G_s \label{eq:1f}\\
  \sum_{i \in V}\ B_{i} \leq& b \label{eq:1g}\\
  B_{i},R_{i,g} \in & \{ 0,1 \} \label{eq:1h}\\
  R_{DA}, y_{DA} =& 1 \label{eq:1i}
\end{align}
\end{subequations}

\subsection{Proof of Proposition 5.1}

\begin{proof} The inequality could be prove with the following step. For every set of graph $X_i$, it is true that: 

\begin{eqnarray}\label{eq:vcg}
\min OBJ_t(X_i) \leq OBJ_t(X_i) \nonumber
\end{eqnarray}

Consequently, we have:
% \begin{proposition}
% \begin{eqnarray}\label{eq:8}
% \displaystyle\sum\limits_{X_i \in X} \min OBJ_t(X_i) \leq \min\limits OBJ_m(G_s)
% \end{eqnarray}
% \end{proposition}
\begin{eqnarray}\label{eq:vcg}
\displaystyle\sum\limits_{X_i \in X} \min OBJ_t(X_i) \leq \displaystyle\sum\limits_{X_i \in X} OBJ_t(X_i) = OBJ_m(G_s) \nonumber
\end{eqnarray}

Applying \textit{min} on both hand side, we obtained inequality (\ref{eq:8}). 
\begin{eqnarray}\label{eq:8}
\displaystyle\sum\limits_{X_i \in X} \min OBJ_t(X_i) \leq \min\limits OBJ_m(G_s)
\end{eqnarray}
Note that optimizing $OBJ_m(G_s)$ actually give us the optimal policy.  Batching solution actually give us the lower bound of the optimal point: 
\begin{eqnarray}\label{eq:8}
\displaystyle\sum\limits_{X_i \in X} \min OBJ_t(X_i) \leq OPT
\end{eqnarray}
\end{proof}
\subsubsection{Number of Sample for Monte Carlo Simulation}
To determine how many graph sample is enough to guarantee that our evaluation is good enough, we use the Hoeffding's Inequality Bound \cite{hoeffding1994probability}. Let us define $X_0, ..., X_n$ is the independence random variable such that $a_i \leq X_i \leq b_i$. Let us say $\epsilon > 0$ is a error of the estimated mean to the true mean. We have the following Hoeffding's Inequality:

\begin{eqnarray}\label{eq:hoefdori}
\Pr(|\hat{X} - \mu| \geq \epsilon) \leq 2e^{\frac{-2n^2\epsilon^2}{\sum_{i=1}^{n} (b_i-a_i)^2}} \leq \alpha
\end{eqnarray}

Where $\hat{X}$ is the sample mean and $\hat{X_n} = \frac{\sum_i X_i}{n}$. $\mu$ is the true mean, n is the sample size and $\alpha$ is the level of significance (upper bound of the error probability). Now let us say we have a blocking strategy $B$ and applying it to sample $G_s$ with where number of sample $n = |G_s|$. Apply $B$ on graph $g_i \in G_s$ yield attacker success score $X^{B}_i$. $X^{B}_i$ is bounded by $0 \leq X^{B}_i \leq 1$ for any i in $G_s$. The sample mean and true mean  of the simulation is $\hat{X^B}$ and $\mu^B$, respectively. We could re-write the inequality as:

\begin{eqnarray}\label{eq:hoefeval}
\Pr(|\hat{X^B} - \mu^B| \geq \epsilon) \leq 2e^{-2n\epsilon ^2} \leq \alpha
\end{eqnarray}

Solving \ref{eq:hoefeval} give use the number of sample required for ensure $\hat{X^B} = \mu^B \pm \epsilon$ with confidence level of $(1-\alpha)$:
\begin{eqnarray}\label{eq:sample}
n \geq \frac{\ln(\frac{2}{\alpha})}{2\epsilon^2}
\end{eqnarray}

In our Monte Carlo Simulation, by testing our blocking polity on $n =$ 100,000 graph samples, our result can be at least 99\% confident ($\alpha = 0.01$) with error within $\epsilon = 5.147\times 10^{-3}$ of the true mean ($\hat{X^B} = \mu^B \pm 5.147\times 10^{-3}$)
\subsection{Clustering based heuristic algorithm}

In this section, we provide the pseudocode for the Clustering algorithm for MIP(m). From line 1 to 4, we extract the feature for each graph using staticMIP (or MIP(m) with m = 1). In line 5, we run clustering algorithm, in our experiment, we use k-mean as our clustering algorithm. We then obtained the centroid coordinate for every cluster and their size. From line 6 to 9, we find the Euclidean distance from a graph's feature to every centroid. The idea is that we will prefer graph which have the closest distance to centroids first as the representative graph for our MIP(m). From line 10 to 15, to avoid the bias to "trivial" graph sample causing by the difference in the size of clusters. In each iteration of $m$, we choose a cluster $c$ where each cluster $c$ with size $size_c$ have a probability of being choose for MIP(m) training of $\frac{size_c}{|G_s|}$. We then pick a graph which have the closest distance to centroid $c$ to be the \textit{representative graph}. The chose graph will be removed from the sample to avoid duplication. 

\begin{algorithm}[!ht]
 \caption{Clustering based heuristic for MIP(m)}
 \label{alg:kmean}
 \begin{algorithmic}[1]
 \renewcommand{\algorithmicrequire}{\textbf{Input:}}
 \REQUIRE graph sample $G_s$, number of cluster $k$, number of graph in MIP(m) $m$, defensive budget $B$\\
 \renewcommand{\algorithmicrequire}{\textbf{Output:}}
 \REQUIRE blocking strategy $B^*$ for graph sample $G_s$ \\
 % \STATE initialize list of feature $feature$
 %  \STATE initialize list of distance to centroid $distance$
 %   \STATE initialize list of graph list for MIP $G$
 \STATE \textbf{for} $g \in G_s$ \textbf{do}
 \STATE \quad $B\ {\leftarrow} \textit{staticMIP}(g)$
 \STATE \quad $feature_{g}\ {\leftarrow} \textit{getGraphFeature} (g, B)$
  \STATE \textbf{end for}
 \STATE $centroids, clusters\_size\ {\leftarrow} \textit{clusteringAlgorithm} (feature, k)$
 
 % // We will find the Euclidean distance from every graph's feature to every centroids
 \STATE \textbf{for} $c$ in $centroids$ \textbf{do}
 \STATE \quad $distance_c\ {\leftarrow} \textit{distanceToCentroid} (feature, c)$
 \STATE \quad Sort $distance_c$ in ascending order
\STATE \textbf{end for}
 \STATE \textbf{for} $i$ in $m$ \textbf{do}
 \STATE \quad Pick a cluster $c$ randomly with distribution based on $clusters\_size$ 
 \STATE \quad $G_i {\leftarrow} \textit{graphWithMinDistance} (G_s, distance_c)$ 
 \STATE \quad Delete $G_i$ from $G_s$
 \STATE \quad Delete $distance_i$ from $distance_c$ 
 \STATE \textbf{end for}
 \STATE $B^*  {\leftarrow} \textit{MIP} (G, m)$
 \STATE \textbf{return} $B^*$ 
 \end{algorithmic}
\end{algorithm}
% \subsection{}

\subsection{Voting based heuristic algorithm}

\begin{algorithm}[H]
 \caption{Voting based heuristic for MIP(m)}
 \label{alg:kmean}
 \begin{algorithmic}[1]
 \renewcommand{\algorithmicrequire}{\textbf{Input:}}
 \REQUIRE graph sample $G_s$, partition size $x$, number of graph in MIP(m) $m$, defensive budget $b$\\
 \renewcommand{\algorithmicrequire}{\textbf{Output:}}
 \REQUIRE blocking strategy $B^*$ for graph sample $G_s$ \\
 % \STATE initialize list of feature $feature$
 %  \STATE initialize list of distance to centroid $distance$
 %   \STATE initialize list of graph list for MIP $G$
 \STATE initialize graph batches X
 \STATE initialize list of blocking strategy B
 \STATE initialize count list C
 \STATE $X \ {\leftarrow} \textit{partitionGraphSample}(G_s, x)$
 \STATE \textbf{for} $x \in X$ \textbf{do}
 \STATE \quad $B_i\ {\leftarrow} \textit{MIP}(x, m)$
 \STATE \textbf{end for}
 \STATE \textbf{for} node \textit{i} in $N$ \textbf{do}
 \STATE \quad \textbf{for} $B_i$ in $B$ \textbf{do}
 \STATE \quad \quad \textbf{if} $i$ in $B_i$ \textbf{do}
 \STATE \quad \quad \quad $C_i \ {\leftarrow} C_i + 1$
 \STATE \quad \textbf{end for}
 \STATE \textbf{end for}
 \STATE Sort $C$ in descending order
 \STATE $B^* \ {\leftarrow} \textit{getTopKNode}(C, b)$
 \STATE \textbf{return} $B^*$ 
 \end{algorithmic}
\end{algorithm}

\subsection{Graph's statistic}

\begin{table}[H]
\begin{center}
\caption{This table provides the size of each graph in terms of the number of nodes, edges, user nodes, and computer nodes.}\label{tab:stat}
\smallskip\noindent
\resizebox{\linewidth}{!}{%
\begin{tabular}{|l||c||c||c||c|}

\hline
       & \textbf{Node}   & \textbf{Edge}    & \textbf{User } & \textbf{Computer} \\
       \hline
\textbf{R2000 } & 5997   & 18795   & 2000  & 2001 \\
\textbf{R4000}  & 12001  & 45780   & 4000  & 4001 \\
\textbf{ADX05 } & 1624   & 6955    & 482   & 526  \\
\textbf{ADX10}  & 3123   & 15767   & 972   & 1026 \\
\textbf{ADU10}  & 13834  & 94878   & 6372  & 366  \\
\textbf{ADU100} & 137315 & 1490766 & 63172 & 3378 \\
       \hline
\end{tabular}}
\end{center}
\end{table}

% \begin{table*}[t]
% \centering
%   \caption{Number of nodes and edges of graphs}
%   \label{tab:locations}
%   \begin{tabular}{rllll}\toprule

%     \bottomrule
%   \end{tabular}
% \end{table*}
% \end{comment}

\end{document}